\documentclass[sigconf]{acmart}

\AtBeginDocument{%
  }

\copyrightyear{2026}
\acmYear{2026}
\setcopyright{cc}
\setcctype{by}
\acmConference[KDD '26]{Proceedings of the 32nd ACM SIGKDD Conference on Knowledge Discovery and Data Mining V.1}{August 09--13, 2026}{Jeju Island, Republic of Korea}
\acmBooktitle{Proceedings of the 32nd ACM SIGKDD Conference on Knowledge Discovery and Data Mining V.1 (KDD '26), August 09--13, 2026, Jeju Island, Republic of Korea}
\acmPrice{}
\acmDOI{10.1145/3770854.3780213}
\acmISBN{979-8-4007-2258-5/2026/08}

\setcopyright{none}
\renewcommand\footnotetextcopyrightpermission[1]{}
\usepackage[english]{babel}
\usepackage{graphicx}
\usepackage{bm} 
\usepackage{booktabs}
\usepackage{amsmath}
\usepackage{tcolorbox}

\usepackage{multirow}
\usepackage{makecell}
\usepackage{microtype}
\usepackage{float}
\usepackage{algorithmic}
\usepackage{algorithm}

\usepackage{amsthm}
\usepackage{xspace}
\usepackage{cleveref}
\usepackage[table,xcdraw]{xcolor}
\usepackage{colortbl}
\usepackage{pifont}
\usepackage{balance}
\usepackage{enumitem}
\usepackage{setspace}

\usepackage{xcolor} %

\definecolor{occolor}{HTML}{f19d38} %
\newcommand{\oc}[1]{{\color{occolor}#1}} %

\newcommand{\bc}[1]{\textcolor{blue}{#1}}
\definecolor{DeepGreen}{RGB}{0, 160, 0} 

\newcommand{\M}{ALARM\xspace}

\newcommand{\NumBaseline}{13\xspace}
\usepackage{subcaption}

\begin{document}

\title[Towards Label-Free Harmful Meme Detection via LMM Agent Self-Improvement]{From Shallow Humor to Metaphor: Towards Label-Free Harmful Meme Detection via LMM Agent Self-Improvement}

\author{Jian Lang}
\email{jian_lang@std.uestc.edu.cn}
\orcid{0009-0009-0876-0497}
\affiliation{%
  \institution{University of Electronic Science and Technology of China}
  \city{Chengdu}
  \state{Sichuan}
  \country{China}
}

\author{Rongpei Hong}
\email{rongpei.hong@std.uestc.edu.cn}
\orcid{0009-0007-4977-1657}
\affiliation{%
  \institution{University of Electronic Science and Technology of China}
  \city{Chengdu}
  \state{Sichuan}
  \country{China}
}

\author{Ting Zhong}
\email{zhongting@uestc.edu.cn}
\orcid{0000-0002-8163-3146}
\affiliation{%
  \institution{University of Electronic Science and Technology of China}
  \city{Chengdu}
  \state{Sichuan}
  \country{China}}

\author{Leiting Chen}
\orcid{0000-0003-2045-7369}
\email{richardchen@uestc.edu.cn}
\affiliation{%
  \institution{University of Electronic Science and Technology of China}
  \city{Chengdu}
  \state{Sichuan}
  \country{China}}
  
\author{Qiang Gao}
\orcid{0000-0002-9621-5414}
\email{qianggao@swufe.edu.cn}
\affiliation{%
  \institution{Southwestern University of Finance and Economics}
  \city{Chengdu}
  \state{Sichuan}
  \country{China}}

\author{Fan Zhou}
\authornote{Corresponding author.}
\email{fan.zhou@uestc.edu.cn}
\orcid{0000-0002-8038-8150}
\affiliation{%
  \institution{University of Electronic Science and Technology of China}
  \city{Chengdu}
  \state{Sichuan}
  \country{China}
}
\affiliation{%
  \institution{Intelligent Digital Media Technology Key Laboratory of Sichuan Province}
  \city{Chengdu}
  \state{Sichuan}
  \country{China}
}

\renewcommand{\shortauthors}{Jian Lang et al.}

\begin{abstract}
The proliferation of harmful memes on online media poses significant risks to public health and stability. 
Existing detection methods heavily rely on large-scale labeled data for training, which necessitates substantial manual annotation efforts and limits their adaptability to the continually evolving nature of harmful content.
To address these challenges,
we present \textbf{\M}, the first l\underline{\textbf{A}}be\underline{\textbf{L}}-free h\underline{\textbf{AR}}mful \underline{\textbf{M}}eme detection framework powered by Large Multimodal Model (LMM) agent self-improvement. 
The core innovation of \M lies in exploiting the expressive information from ``shallow'' memes to iteratively enhance its ability to tackle more complex and subtle ones.
\M consists of a novel \textit{Confidence-based Explicit Meme Identification mechanism} that isolates the explicit memes from the original dataset and assigns them pseudo-labels.
Besides, a new \textit{Pairwise Learning Guided Agent Self-Improvement paradigm} is introduced, where the explicit memes are reorganized into contrastive pairs (positive \textit{vs.} negative) to refine a learner LMM agent. 
This agent autonomously derives high-level detection cues from these pairs, which in turn empower the agent itself to handle complex and challenging memes effectively.
Experiments on three diverse datasets demonstrate the superior performance and strong adaptability of \M to newly evolved memes.
Notably, our method even outperforms label-driven methods. 
These results highlight the potential of label-free frameworks as a scalable and promising solution for adapting to novel forms and topics of harmful memes in dynamic online environments.
\end{abstract}




\maketitle

\section{Introduction}
\label{sec:introduction}

The proliferation of multimedia ushered in a novel multimodal form of communication: \textit{memes}, which blend images with concise textual elements to deliver humor, satire, or commentary~\cite{yang2023invariant, prajwal2019towards}.
However, 
memes have increasingly been exploited as vehicles for spreading harmful content across online social media. 
These harmful memes often attack individuals or communities with respect to their race, gender, nationality, etc., fostering serious discrimination and conflict~\cite{lee2021disentangling}.
Consequently, developing effective approaches to identify and counteract harmful memes has become a pressing issue  with important real-world significance.

\textit{However}, accurately detecting the harmful memes is difficult due to the subtle harmful expression and the fast-evolving content. 
Early methods mainly leveraged the pre-trained Vision-Language Models (VLMs)~\cite{radford2021learning} to learn the cross-modal correlations and excavate deep harmful content~\cite{kiela2020hateful, pramanick2021momenta}. 
They either fine-tuned the VLMs or combined the frozen VLMs with learnable classifiers to detect the underlying harmful content~\cite{cao2022prompting, cao2023procap}.
The prevalence of Large Multimodal Models (LMMs)~\cite{achiam2023gpt, han2024multimodal} has enabled zero-shot harmful meme detection in recent studies~\cite{huang2024towards, lin2024towards}. 
\textit{Nevertheless}, when confronted with more nuanced and complex harmful forms, LMMs often rely on task-specific fine-tuning with extensive labeled meme data to achieve satisfactory performance~\cite{lin2023beneath, ji2023identifying}.
Despite the progress, significant unresolved challenges remain:

\begin{figure}[t]
    \centering
    \includegraphics[width=\columnwidth]{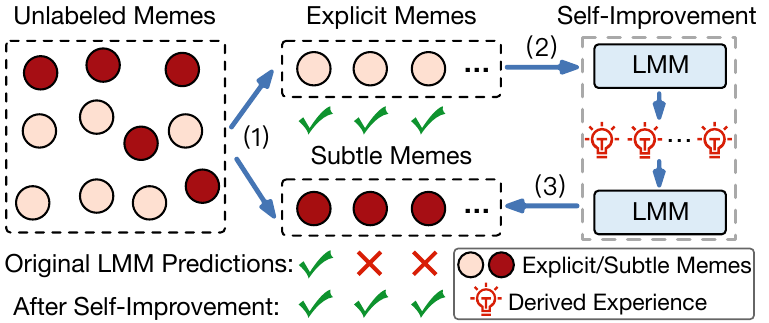}
    \vspace{-7mm}
    \caption{Concept diagram of our framework \M, where the LMM agent achieves self-improvement by distilling the ``powerful experiences'' from the explicit memes, therefore delivering enhanced detections on the subtle ones.  
    }
    \label{fig:intro}
    \vspace{-4mm}
\end{figure}

\noindent \textbf{Challenge 1: Requirement for Large Amount of Annotated Data.} 
Existing methods heavily depend on large volumes of annotated data to enable supervised model training or fine-tuning~\cite{pramanick2021momenta, cao2022prompting, cao2023procap}. 
\textit{However}, labeling large volume of memes requires both labor-intensive and domain expertise, and may encounter challenges such as copyright restriction~\cite{hee2024bridging}. 
As a result, the scarcity of annotated data restrict the performance and generalizability of existing label-driven methods, as they are limited to learning from a narrow and incomplete spectrum of harmful meme forms.

\noindent \textbf{Challenge 2: Difficulty of Adapting to the Fast-Evolving Harmful Content.}
The nature of harmful content and its manifestations in online media evolves rapidly, driven by breaking news or emerging events~\cite{cao2024modularized, huang2024towards, lang2025biting}. 
Existing label-driven methods struggle to address these newly emerging memes, as such content are often out-of-distribution compared to previously seen data. 
Moreover, as highlighted in Challenge 1, annotating first-appearing memes for training requires substantial time due to the complexity of labeling, which inherently conflicts with the timeliness of harmful content.
\textit{Consequently}, developing more responsive detection approaches that can tackle new and unseen harmful memes without relying on extensive annotated data remains an open and non-trivial challenge.

Intuitively, to achieve label-free detection, the key lies in maximizing the utilization of powerful information available in unannotated data. 
For harmful memes, we observe that: 
\textbf{(O1)} On one hand, for a specific harmful topic, some memes \textit{explicitly} express harmful intent, making them relatively easy to detect, while others convey harm in a more \textit{subtle} or \textit{implicit} manner, posing greater challenges for detection. 
As shown in \Cref{fig:ALARM} (1), the left meme explicitly conveys anti-Muslim discrimination, whereas the right meme embeds the same harmful intent in a more subtle manner --- associating a pregnant woman with the caption ``bomb''.
\textit{However}, if the model first gathers ``experiences'' from the explicit discrimination in the left meme --- dehumanizing a Muslim child by equating them to a bomb --- it can effectively help the model to identify the subtle harmful message in the right meme.
\textbf{(O2)} On the other hand, within newly emerging harmful memes on trending topics, there are also a portion of memes that exhibit explicit harmful expressions.
These explicit memes have the potential to provide valuable ``experiences'' for detecting the more challenging memes in such topics.
In light of these observations, a natural question arises: 
\textit{``Can we accumulate diverse and valuable references from unlabeled easy-to-detect harmful memes to aid in the detection of more nuanced and challenging cases, thereby eliminating the need for any annotated data?''}

We affirmatively answer this question by proposing \textbf{\M}, the first l\underline{\textbf{A}}be\underline{\textbf{L}}-free h\underline{\textbf{AR}}mful \underline{\textbf{M}}eme detection framework powered by LMM agent self-improvement. 
Unlike previous methods that heavily rely on large-scale annotated datasets and gradient-based supervised learning, our framework marks a significant advancement in achieving an entirely label-free detection solution: It requires no model-weights optimization and no supervised training, allowing \M to adapt flexibly and immediately to newly emerging forms or topics of harmful memes without any additional training effort.
The core idea of this framework is inspired by the remarkable human learning process: 
Humans build expertise by starting with simpler cases, accumulating and distilling experiences from these cases, and then applying the learned knowledge to tackle more complex situations~\cite{thrun1998learning}.
Similarly, \M equips the learner LMM agent with the capability to autonomously harvest diverse experiences and informative references from easy-to-detect memes. 
These experiences and references function as implicit ``gradient signals,'' enabling the LMM agent to continually refine itself without any additional training to detect the increasingly subtle and challenging harmful memes. 
A concise conceptual illustration of our proposed \M is provided in~\Cref{fig:intro}.

Specifically, we first propose a novel \textit{Confidence-based Explicit Meme Identification} mechanism, which facilitates the LMM to isolate the explicit memes through a prediction probabilities based ordering and assigns pseudo-labels to high-confidence memes.
Besides, inspired by the compelling design of triple contrastive learning that improves model performance by capturing nuanced differences between similar positive and negative samples~\cite{yang2022vision}, we introduce a new \textit{Pairwise Learning Guided Agent Self-Improvement} paradigm.
In this paradigm, explicit memes are organized into contrastive pairs using multimodal retrieval, where each pair comprises a pseudo-harmful meme and a 
closely related pseudo-benign one. 
Subsequently, the LMM is positioned as a self-driven learner agent, dedicated to analyzing and summarizing the fine-grained differences between the pseudo-harmful and -benign memes in each pair. 
This process allows the LMM agent to distill meaningful, diverse, and fine-grained experiences by focusing on subtle and nuanced discrepancies.
Finally, building on the acquired experiences, the LMM agent autonomously purifies a set of high-level and well-generalized detection references. 
These self-generated references enable the agent itself to effectively identify and tackle newly emerging and more challenging harmful memes, ensuring robust detection in dynamic and fast-evolving conditions.
To sum up, our main contributions are summarized as follows:

\begin{itemize}[leftmargin=*,
                topsep=0pt plus 0.5pt,
                partopsep=0pt plus 0.5pt,
                itemsep=0pt plus 0.5pt,
                parsep=0pt plus 0.5pt]
\item 
We present the first method \M to effectively harness the expressive potential of the unlabeled explicit samples for tackling complex ones, achieving promising label-free harmful meme detection through an innovative LMM agent self-improvement framework.
Moreover, these label- and training-free merits enable \M to rapidly adapt to newly emerging forms and topics of harmful memes, while without annotating even one sample.

\item We introduce a new Confidence-based Explicit Meme Identification mechanism, which utilizes LMMs to 
facilitate the isolation of the explicit memes 
and endow these memes with pseudo-labels. 

\item We propose a fresh Pairwise Learning-Guided Agent Self-Improve-ment paradigm, which builds contrastive pairs from explicit memes and facilitates the LMM agent to autonomously refine valuable references from these pairs for better detection.
\end{itemize}

Comprehensive experiments on diverse benchmarks demonstrate the efficacy of our \M. 
Notably, \M not only outperforms the competitive label-driven detection methods but also showcases strong adaptability to newly emerging memes, highlighting its capability as a timely, robust, and scalable label-free detection solution.
The code and data required to reproduce our results are available at \bc{\url{https://github.com/Jian-Lang/ALARM}}.

\section{Related Work}
\subsection{Harmful Meme Detection}
Harmful meme detection aims to judge whether memes contain harmful content by analyzing the meme images and textual elements. 
Early works primarily adopted various cross-modal learning methods to perform the prediction~\cite{kiela2020hateful, pramanick2021detecting, pramanick2021momenta}. 
To excavate deep harmful meanings in memes, some researchers performed task-specific fine-tuning on the pre-trained VLMs or added learnable classifiers behind the frozen VLMs to detect the harmful memes~\cite{cao2022prompting, cao2023procap}.
With the popularity of LMMs~\cite{achiam2023gpt}, recent work have turned to incorporating LMMs in detection with zero-shot manner and further fine-tuned them for better prediction~\cite{alayrac2022flamingo, lin2023beneath, ji2023identifying, tian2024learning}.
Nevertheless, all these label-driven approaches heavily rely on large amount of annotated memes, which are challenging to guarantee and hinder their ability to adapt to fast-evolving harmful content.

Currently, some studies leveraged few-shot learning method on LMMs to achieve a relative low-resource regime, alleviating the requirement of labeled data~\cite{hee2024bridging, cao2024modularized, huang2024towards}. 
However, they still require a few annotated memes and failed to utilize the expressive information from unlabeled data, incurring compromised performance and limited generalizability to newly emerging memes.
A recent work, LoREHM~\cite{huang2024towards}, also explored leveraging LMM agents for low-resource meme detection. 
\textit{However}, it still relies on a small number of annotated memes and extracts information only from these limited samples, failing to reveal the key insight for achieving label-free detection: fully exploiting abundant unlabeled data. 
As a result, its performance and effectiveness remains constrained.
\textit{In contrast}, \M proposes to excavate valuable information from unlabeled explicit memes and leverage these knowledge to achieve ``self-improvement'', significantly improving the detection ability on more challenging memes under label-free condition.

\subsection{Agent Self-Improvement} 
Machine learning has long explored the potential of autonomous agents~\cite{dorigo1994robot, langley2017explainable}. 
Recently, the integration of LMMs into these agents has garnered significant attention in both research and industry communities~\cite{liu2025visualagentbench}. 
To address the substantial computational demands and the reliance on extensive labeled datasets inherent to traditional reinforcement learning-based agent optimization~\cite{xu2020multi, abramson2022improving}, 
recent research has proposed an alternative efficient \textit{agent self-improvement} paradigm. 
This paradigm leverages LMMs as agents to generate feedback itself and progressively refine their outputs based on this self-generated feedback, eliminating the requirement of extensive annotated data for supervised learning~\cite{madaan2023selfrefine, huang2023large, zhao2024expel}.
In this work, we draw inspiration from the unsupervised and evolutionary nature of this strategy and propose a novel label-free agentic framework. 
It introduces an LMM agent to derive high-level references from unlabeled shallow samples and regards the self-generated references as ``verbal gradient'' to refine and optimize performance of the agent itself on more challenging samples.

\section{Methodology}
\subsection{Preliminary}
\noindent \textbf{Problem Statement.}
Given a set of $N$ memes $\mathcal{D} = \{\mathcal{M}_i\}_{i=1}^N$, the harmful meme detection is to determine whether each meme $\mathcal{M}_i$ is \textit{harmful} or \textit{benign} based on its meme image $\mathcal{I}_i$ and textual element (i.e., text embedded on the meme image) $\mathcal{T}_i$.
Existing label-driven methods first optimize their models with the annotated training set $\mathcal{D}_\text{train}$ and then evaluate the models on the test set $\mathcal{D}_\text{test}$.

\noindent \textbf{Our Pipeline.}
To tackle labor-intensive nature of labeling memes and the fast-evolving harmful content, we redefine the harmful meme detection to a label-free task. 
Our core idea is to acquire valuable detection references from unlabeled easy-to-detect memes and leverage these references to guide the detection of difficult cases.
Based on this, we first split the original meme set $\mathcal{D}$ into an explicit set $\mathcal{D}_{\text{exp}}$ and a subtle set $\mathcal{D}_{\text{sub}}$.
Subsequently, we propose an LMM agent $\mathcal{A}_\text{LMM}$ to acquire abundant case-based experiences $\mathcal{E}$ from $\mathcal{D}_\text{exp}$.
The LMM agent is then instructed to iteratively purify a set of high-level detection references $\mathcal{R}$ by meticulously analyzing each experience $e \in \mathcal{E}$.
These references finally serve to enhance the agent's detection capability on the challenging subset $\mathcal{D}_{\text{sub}}$.

\begin{figure*}[t]
  \centering
  \includegraphics[width=\textwidth]{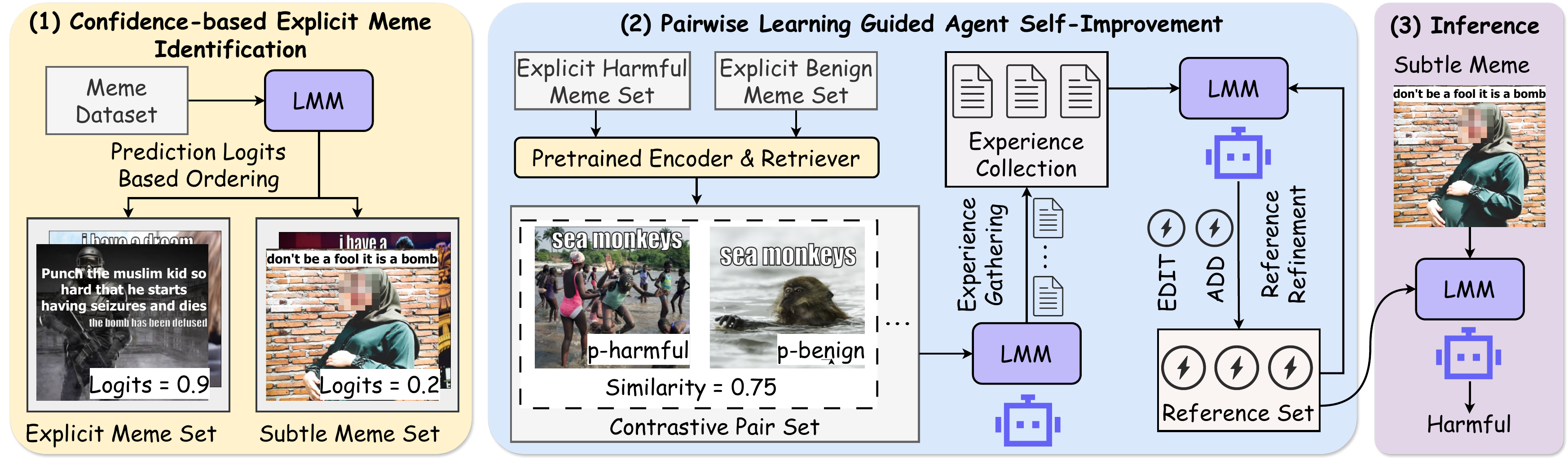}
  \vspace{-6mm}
  \caption{The overall framework of \M. (1) The original meme dataset are divided into both explicit and subtle subsets through a Confidence-based Explicit Meme Identification mechanism. (2) The pseudo-labeled explicit memes in two categories are reorganized into contrastive pairs. 
  And a self-driven LMM agent is positioned to gather experiences and derive references from these pairs via a Pairwise Learning Guided Agent Self-Improvement paradigm.
  (3) At inference, the self-generated references are leveraged to help LMM agent itself in tackling the subtle and challenge meme.}
  \label{fig:ALARM}
   \vspace{-3mm}
\end{figure*}

\noindent \textbf{Structure of Methodology.} \Cref{fig:ALARM} demonstrates the core components and their relationships in \M. 
The following sections delve into the details of each component: 
\Cref{method:pseudo_label} introduces the identification of explicit memes.
Subsequently, these explicit memes are fully exploited to realize the self-improvement of LMM agent in \Cref{method:self_improvement}. 
The last section \Cref{method:inference} presents the inference process of our \M on more challenging memes.

\subsection{Confidence-based Explicit Meme Identification}
\label{method:pseudo_label}
In the context of memes, while some memes possess subtle and challenging meanings, others exhibit explicit forms and are relatively easy-to-detect.
With the rise of LMMs, zero-shot detection on these explicit memes with LMMs has become plausible. 
As statistical models, the prediction probabilities (probs) generated by LMMs inherently reflect a measure of confidence in their outputs~\cite{stolfo2024confidence}.
In light of these observations and prior findings~\cite{liang2020we, wang2021tent}, we posit that memes with high prediction probs (confidence) from LMMs are likely to contain explicit content and are more likely to be correctly identified by the LMMs.
To validate this, we conducted an experiment to examine the relationship between LMM's prediction confidence and accuracy, and results further support our hypothesis (cf. \Cref{subsec:pre-exp} for evidence).
To this end, we propose a novel \textit{Confidence-based Explicit Meme Identification mechanism}, which splits the original meme set $\mathcal{D}$ into both an explicit set $\mathcal{D}_{\text{exp}}$ and a subtle set $\mathcal{D}_{\text{sub}}$ and endow the memes in $\mathcal{D}_{\text{exp}}$ with pseudo-labels.

Specifically, for each meme $\mathcal{M}_i \in \mathcal{D}$, we first formulate a zero-shot detection instruction for the LMM. The meme's image $\mathcal{I}_i$ and textual element $\mathcal{T}_i$ are then provided as input to the LMM to yield the prediction probs:
\begin{equation}
\label{eq:1}
    \boldsymbol{z}_i = \mathcal{F}_\text{LMM}(\mathcal{P}_\text{ide}, \mathcal{I}_i, \mathcal{T}_i),
\end{equation}
where $\mathcal{F}_\text{LMM}$ is the decoder of LMM, $\boldsymbol{z}_i = [z_{\text{hm}, i}, z_{\text{bn}, i}] \in \mathbb{R}^{2}$ denotes the prediction probs for \underline{h}ar\underline{m}ful or \underline{b}e\underline{n}ign\footnote{In the implementation, we guide the LMM to produce ``1'' to represent ``harmful'' and ``0'' to represent ``benign'' for the generation of prediction probs.}, and $\mathcal{P}_\text{ide}$ is the prompt for identifying explicit memes with a concise version shown as: 

\textit{``Given a meme with its image [$\mathcal{I}_i$] and textual element [$\mathcal{T}_i$], determine whether it contains harmful content. If the meme promotes harmful, classify it as `harmful.' Otherwise, classify it as `benign'...""}

We then order the memes in $\mathcal{D}$ based on the prediction probs and isolate high confidence memes to form the explicit set $\mathcal{D}_{\text{exp}}$:
\begin{equation}
\label{eq:2}
       \mathcal{D}_{\text{exp}} = \{ \mathcal{M}_i \mid \mathcal{M}_i\in \mathcal{D},\,  \max(z_{\text{hm}, i}, z_{\text{bn}, i}) > Q_{1-\tau\%} \}, 
\end{equation}
where $Q_{1-\tau\%}$ is the quantile-based threshold to ensure selection of memes with the top-$\tau\%$ prediction probs. 
The complement set of $\mathcal{D}_{\text{exp}}$ in $\mathcal{D}$ is marked as subtle set: $\mathcal{D}_\text{sub} = \mathcal{D} \backslash \mathcal{D}_{\text{exp}}$. For each meme $\mathcal{M}_i$ in $\mathcal{D}_{\text{exp}}$, we yield the pseudo-labels following the rules:
\begin{equation}
\label{eq:3}
    \text{Pseudo-label}(\mathcal{M}_i) =
\begin{cases}
\text{harmful}, & \text{if } z_{\text{hm}, i} \geq z_{\text{bn}, i}; \\
\text{benign}, & \text{if } z_{\text{hm}, i} < z_{\text{bn}, i}.
\end{cases}
\end{equation}
The $\mathcal{D}_{\text{exp}}$ will be fully harnessed, yielding the powerful detection knowledge for identifying more challenging memes in $\mathcal{D}_{\text{sub}}$.

\subsection{Pairwise Learning Guided Agent Self-Improvement}
\label{method:self_improvement}
Inspired by the effective triple contrastive learning approaches~\cite{yang2022vision}, which significantly improve the discriminative capability of models in classification by providing models with similar positive and negative samples, we introduce a fresh \textit{Pairwise Learning Guided Agent Self-Improvement paradigm}.
This paradigm first reorganizes the explicit memes in $\mathcal{D}_{\text{exp}}$ into contrastive pairs, where each pair consists of one pseudo-harmful meme with a semantically similar pseudo-benign one.
An LMM is then positioned as a self-driven learner agent to gather diverse case-based experiences from the contrastive pairs and refine high-level detection references from the acquired experiences for identifying more challenging memes.

\subsubsection{Multimodal Retrieval-based Contrastive Pairing}
To construct contrastive pairs, we begin by dividing the explicit meme set into two disjoint subsets based on the pseudo-category of the memes:
\begin{align}
\label{eq:4}
\mathcal{D}_{\text{exp}}^{\text{hm}} &= \{\mathcal{M}_i \mid \mathcal{M}_i \in \mathcal{D}_\text{exp},\, \text{category}(\mathcal{M}_i) = \text{harmful}\},\\
\label{eq:5}
\mathcal{D}_{\text{exp}}^{\text{bn}} &= \{\mathcal{M}_i \mid \mathcal{M}_i \in \mathcal{D}_\text{exp},\, \text{category}(\mathcal{M}_i) = \text{benign}\},
\end{align}
where $\mathcal{D}_{\text{exp}} = \mathcal{D}_{\text{exp}}^{\text{hm}} \cup \mathcal{D}_{\text{exp}}^{\text{bn}}$. 
Subsequently, for each pseudo-harmful meme $\mathcal{M}_i \in \mathcal{D}_{\text{exp}}^{\text{hm}}$, we employ a similarity-based multimodal retrieval to identify its most semantically similar pseudo-benign counterpart in $\mathcal{D}_{\text{exp}}^{\text{bn}}$ and form a contrastive pair set: 
\begin{equation}
\label{eq:6}
    \begin{aligned}
        \mathcal{C}_{\text{hm} \to \text{bn}} = \big\{ (\mathcal{M}_i, \mathcal{M}_j^*) \mid \mathcal{M}_i \in \mathcal{D}_{\text{exp}}^{\text{hm}}, \,
        \mathcal{M}_j^* = \\ \arg\max\limits_{\mathcal{M}_j \in \mathcal{D}_{\text{exp}}^{\text{bn}}} \text{sim}(\mathcal{M}_i, \mathcal{M}_j) \big\},
    \end{aligned}
\end{equation}
where $\mathcal{M}_j^*$ refers to the retrieved pseudo-benign meme that is most semantically related to $\mathcal{M}_i$, and $\text{sim}(\mathcal{M}_i, \mathcal{M}_j)$ is defined as:
\begin{equation}
\label{eq:7}
    \text{sim}(\mathcal{M}_i, \mathcal{M}_j) = \frac{{\Psi_t(\mathcal{T}_i)}^{\top} \Psi_t(\mathcal{T}_j)}{\|\Psi_t(\mathcal{T}_i)\| \|\Psi_t(\mathcal{T}_j)\|} + \frac{{\Psi_v(\mathcal{I}_i)}^{\top} \Psi_v(\mathcal{I}_j)}{\|\Psi_v(\mathcal{I}_i)\| \|\Psi_v(\mathcal{I}_j)\|},
\end{equation}
where $\Psi_t(\cdot)$ and $\Psi_v(\cdot)$ are frozen text and vision encoders from pre-trained models (e.g., CLIP~\cite{radford2021learning}).
Symmetrically, each meme $\mathcal{M}_j \in \mathcal{D}_{\text{exp}}^{\text{bn}}$ is treated as a query to retrieve its most similar pseudo-harmful counterpart from $\mathcal{D}^\text{hm}_\text{exp}$, yielding another contrastive pair set $\mathcal{C}_{\text{bn} \to \text{hm}}$.
Finally, the two sets are merged to form the complete contrastive set $\mathcal{C}$ while ensuring no duplicate pairs.
The contrastive pairs stored in $\mathcal{C}$ will be utilized as a valuable resource exploited from the unlabeled dataset to realize the label-free detection.

\subsubsection{Pairwise Learning-guided Experience Gathering}
Building on the contrastive pair set, we propose an LMM agent, denoted as $\mathcal{A}_\text{LMM}$, to analyze the nuanced differences between the pseudo-positive and pseudo-negative memes within each pair, progressively accumulating a rich repository of fine-grained experiences.
Specifically, we employ a two-step Chain-of-Thought (CoT)~\cite{wei2022chainofthought} prompting strategy to effectively gather experiences from the contrastive pairs while mitigating the hallucination. 
In the first step, the agent is instructed to briefly describe the multimodal content of both memes in each pair. 
In the second step, the agent is prompted to carefully analyze and summarize the discrepancies within each pair, resulting in a collection of case-based experiences:
\begin{equation}
\label{eq:8}
    \mathcal{E} = \{ \mathcal{A}_\text{LMM}(\mathcal{P}_\text{exp}, \mathcal{I}_i, \mathcal{T}_i, \mathcal{I}_j, \mathcal{T}_j), \mid (\mathcal{M}_i,\mathcal{M}_j)\in \mathcal{C}\},
\end{equation}
where $\mathcal{P}_\text{exp}$ denotes the CoT prompt with a brief version shown as:

\textit{``Given the harmful Meme i with its image and textual element [$\mathcal{I}_i$] and [$\mathcal{T}_i$], with a similar but benign Meme j with its image and textual element [$\mathcal{I}_j$] and [$\mathcal{T}_j$], please finish the following two steps...''}

\subsubsection{Reference Refinement}
Directly feeding gathered massive albeit case-specific experiences to the LMMs faces the context-length limitation~\cite{tworkowski2023focused}.
To maximally exploit the abundant and diverse experiences while addressing this limitation, we instruct the agent $\mathcal{A}_\text{LMM}$ to autonomously analyze these experiences and derive high-level detection references 
from two complementary perspectives.
First, the agent should refine discriminative signals that effectively characterize specific and scarce harmful patterns.
Second, the agent should purify universal detection principles aimed at capturing shared and common harmful characteristics.
Specifically, we first initialize an empty reference set: $\mathcal{R}_0 = \emptyset$, and define four atomic agentic operations~\cite{zhao2024expel} that can be performed on the reference set: \textbf{ADD}, to generate a new detection reference with an initial
importance count of two; \textbf{UPVOTE/DOWNVOTE}, to increase/decrease the importance of an existing detection reference\footnote{A detection reference is removed when its importance count is decreased to zero.}; and \textbf{EDIT}, to modify an existing detection reference.
Subsequently, $\mathcal{A}_\text{LMM}$ is equipped with the four operations and iteratively update the reference set by analyzing each experience $e_i \in \mathcal{E}$,
\begin{equation}
\label{eq:9}
    \mathcal{R}_i = \mathcal{A}_\text{LMM}(\mathcal{P}_\text{ref}, \mathcal{R}_{i-1}, e_i), \quad \forall  e_i \in \mathcal{E},
\end{equation}
where $\mathcal{R}_i$ signifies the reference set at $i$-iteration, $\mathcal{P}_\text{ref}$ represents the prompt designed to guide the agent in updating and maintaining the reference set, with a concise version shown as:

\textit{``Given a harmful meme detection reference set [$\mathcal{R}_{i-1}$], and an upcoming case-based experience [$e_i$] that summarizes the key differences between two similar yet distinct categories of memes, please update the reference set by conducting four atomic operations...''}

After iteratively processing the entire set of experiences, the agent $\mathcal{A}_\text{LMM}$ purifies a set of valuable references, denoted as $\mathcal{R}$. 
Notably, the capacity of the reference set $\mathcal{R}$ is constrained to $L$, ensuring both the sufficiency and representativity of the references.

\subsection{Inference}
\label{method:inference}
After deriving a set of valuable inferences, we target the LMM agent $\mathcal{A}_\text{LMM}$ to detect subtle harmful content in more challenging memes within $\mathcal{D}_\text{sub}$ with the guidance of these self-generated references. Specifically, the image $\mathcal{I}_i$, textual element $\mathcal{T}_i$, and the reference set $\mathcal{R}$ are fed to $\mathcal{A}_\text{LMM}$, resulting in the final prediction for each challenging meme $\mathcal{M}_i \in \mathcal{D}_\text{sub}$:
\begin{equation}
\label{eq:10}
    \hat{y}_i = \mathcal{A}_\text{LMM}(\mathcal{P}_\text{inf}, \mathcal{R}, \mathcal{I}_i, \mathcal{T}_i),
\end{equation}
where $\hat{y}_i$ is the predicted category of meme $\mathcal{M}_i$, and $\mathcal{P}_\text{inf}$ denotes the prompt employed to instruct the agent for inference, with a concise version presented below:

\textit{``Given a harmful meme detection reference set [$\mathcal{R}$], and a coming meme with its image [$\mathcal{I}_i$] and textual element [$\mathcal{T}_i$], your task is to determine whether this meme is harmful or not...''}

By equipping the self-generated references distilled from easy-to-detect memes, the LMM agent is self-improved to identify the subtle harmful content in challenging memes. 

\section{Experiments}
\label{sec:experiment}

\subsection{Experimental Settings}
\label{subsec:exp-set}
A concise summary of the experimental settings is presented below, with detailed descriptions regarding datasets, baselines, and implementation available in the \Cref{app-exp-set}.

\begin{table}[t]
  \centering
  \setlength{\tabcolsep}{2.5pt}
  \caption{Statistics of three datasets.}
  \vspace{-3mm}
  \label{tab:dataset}
  \resizebox{\linewidth}{!}{
  \begin{tabular}{@{}ccccccc@{}}
    \toprule
    Dataset  &\# Harmful &\# Benign &\# Total  &\# Train  &\# Test &\# Language \\
    \midrule
    FHM	  & 3,300 & 5,700 & 9,000 & 8,500 & 500 & English \\
    MAMI   & 5,500 & 5,500 & 11,000 & 10,000 & 1,000 & English  \\
    ToxiCN  & 3,827 & 8,173 & 12,000 & 9,600 & 2,400 & Chinese   \\
    \bottomrule
    
  \end{tabular}
  }
  \vspace{-3mm}
\end{table}

\begin{table*}[t]
    \caption{Performance comparison on the FHM, MAMI and ToxiCN datasets. The best results are in \textbf{black bold}, while the second are \underline{underlined}. 
    Higher values of Accuracy and Macro-F1 indicate better performance. 
    $\dagger$ denotes the baselines which also serve as LMM backbones for \M and baseline LoReHM.}
    \label{tab:main-result}
    \vspace{-2mm}
    \centering
    \setlength{\tabcolsep}{5.5pt}
    \resizebox{\linewidth}{!}{
    \begin{tabular}{l|c|c|cccccc}
    \toprule
     \multirow{2}{*}{\textbf{Methods}} & \multirow{2}{*}{\makecell[c]{\textbf{Training} \\ \textbf{Free}}} & \multirow{2}{*}{\textbf{Setting}} & \multicolumn{2}{c}{\textbf{FHM}} & \multicolumn{2}{c}{\textbf{MAMI}} & \multicolumn{2}{c}{\textbf{ToxiCN}} \\ 
     & &  & \textbf{Accuracy} &  \textbf{Macro-F1} &  \textbf{Accuracy} &  \textbf{Macro-F1} & \textbf{Accuracy} &  \textbf{Macro-F1} \\
    \midrule
    MOMENTA & \ding{55} & Label-Driven & 61.34 & 57.45 & 72.10 & 71.26 & \underline{77.87} & 70.25 \\
    PromptHate & \ding{55} & Label-Driven  & 72.20 & 71.83 & 70.40 & 69.92 & 76.04 & 67.45 \\
    MR.HARM & \ding{55} & Label-Driven  & 75.40 & 75.10  & 72.40 & 71.49  & 77.16 & 69.55 \\
    Pro-Cap & \ding{55} & Label-Driven  & 74.95 & 71.68 & 73.06 & 72.40 & 75.70 & \underline{71.36} \\
    ISM & \ding{55} & Label-Driven  & 70.45 & 68.77 & 66.40 & 65.35 & 74.92 & 67.02 \\
    ExplainHM & \ding{55} & Label-Driven & \underline{75.60} & \underline{75.39} & 73.70 & 73.25 & 75.20 & 67.60\\
    HHPrompt & \ding{55} & Label-Driven & 70.40 & 69.01 & 75.30 & 75.10 & 72.29  & 60.03\\
    \midrule
    OPT-30B & \ding{51} & Few Shot & 54.20 & 50.82 & 63.40 & 63.40 & 49.75 & 48.99 \\
    OpenFlamingo-9B & \ding{51} & Few Shot & 51.60 & 51.52 & 52.70 & 46.80 & 49.01 & 48.69 \\
    $\text{Qwen2.5-VL-72B}^{\dagger}$ & \ding{51} & Few Shot & 71.20 & 71.02 & 78.70 & 78.65 & 74.08 & 66.40\\
    $\text{GPT-4o}^{\dagger}$ & \ding{51} & Few Shot & 66.60 & 65.74  & 80.80 & 80.52 & 71.26 & 50.19\\
    Mod-HATE & \ding{55} & Few Shot & 57.60 & 53.88 & 69.05 & 68.78 & 61.24 & 59.71 \\
    LoReHM (Qwen2.5-VL-72B) & \ding{51} & Few Shot & 69.00 & 68.67  & 79.80 & 79.76 & 75.29 & 64.36 \\
    LoReHM (GPT-4o) & \ding{51} & Few Shot & 70.20 & 70.14 & \underline{83.00} & \underline{82.98} & 72.55 & 57.85 \\
    \midrule
    \rowcolor{gray!10}
    \textbf{\M (Qwen2.5-VL-72B)} & \ding{51} & Label-Free & \textbf{75.80} & \textbf{75.79} &  81.28 & 81.25 & \textbf{79.21} & \textbf{72.51} \\ \rowcolor{gray!10}
    \textbf{\M (GPT-4o)} & \ding{51} & Label-Free & 72.80 & 72.75 & \textbf{85.50} & \textbf{85.50} & 77.45 & 67.87 \\
    \bottomrule
    \end{tabular}
    }
    \vspace{-2mm}
\end{table*}

\noindent \textbf{Datasets.}
To comprehensively evaluate the effectiveness and generalizability of our \M, we conduct experiments on three publicly and diverse benchmarks, including two widely adopted English meme datasets FHM~\cite{kiela2020hateful} , MAMI~\cite{fersini2022semeval}, and a Chinese meme dataset ToxiCN~\cite{lu2024towards}. 
Each dataset is split into training and test sets following the original papers, with statistics presented in \Cref{tab:dataset}.

\noindent \textbf{Baselines.}
We compare our \M with \NumBaseline competitive baseline models, which can be divided into two groups: 
(1) \textit{Label-Driven Methods} which require annotated training data for optimizing the detection models, including MOMENTA~\cite{pramanick2021momenta}, PromptHate~\cite{cao2022prompting}, MR.HARM~\cite{lin2023beneath}, Pro-Cap~\cite{cao2023procap}, ISM~\cite{yang2023invariant}, ExplainHM~\cite{lin2024towards}, and HHPrompt~\cite{xu2025hyperhateprompt}.
(2) \textit{Few-Shot Learning Methods} which only leverage a few labeled samples to enhance their detection capability, including OPT-30B~\cite{zhang2022opt}, OpenFlamingo-9B~\cite{alayrac2022flamingo}, Qwen2.5-VL-72B~\cite{bai2025qwen2}, GPT-4o~\cite{openai2024gpt}, Mod-HATE~\cite{cao2024modularized}, and LoReHM~\cite{huang2024towards}.
Notably, Mod-HATE is trained on the labeled few samples, while the other models utilize the few samples as in-context demonstrations.

\noindent \textbf{Metrics.} 
Following prior works in harmful meme detection~\cite{lin2023beneath, cao2024modularized, huang2024towards}, we adopt Accuracy and Macro-F1 as the metrics to evaluate the performance of \M and each baseline models.

\noindent \textbf{Implementation Details.}
We mainly adopt open-source Qwen2.5-VL-72B model as LMM backbone.
Given the high scalability of \M, it can also be implemented with close-source LMMs and we select GPT-4o (gpt-4o-2024-11-20) as another backbone.
However, the Confidence-based Explicit Meme Identification mechanism still requires the open-source model Qwen2.5-VL-72B for execution.
During the multimodal retrieval, we employ text and vision encoders from the pre-trained CLIP model. 
The confidence selection ratio $\tau$ is set to 0.2, and the reference set capacity $L$ is set to 15 across all three datasets.
While \M is capable of directly processing the test set for reference derivation, to ensure a fair comparison, we restrict \M to utilize only the training set for reference generation.
For few-shot learning baselines, we follow the prior work~\cite{huang2024towards} to provide them with 50 examples.
The experiments are conducted on an NVIDIA L40s GPU.

\begin{figure}[t]
    \centering
    \begin{subfigure}[b]{0.491\columnwidth}
    \centering
    \includegraphics[width=\columnwidth]{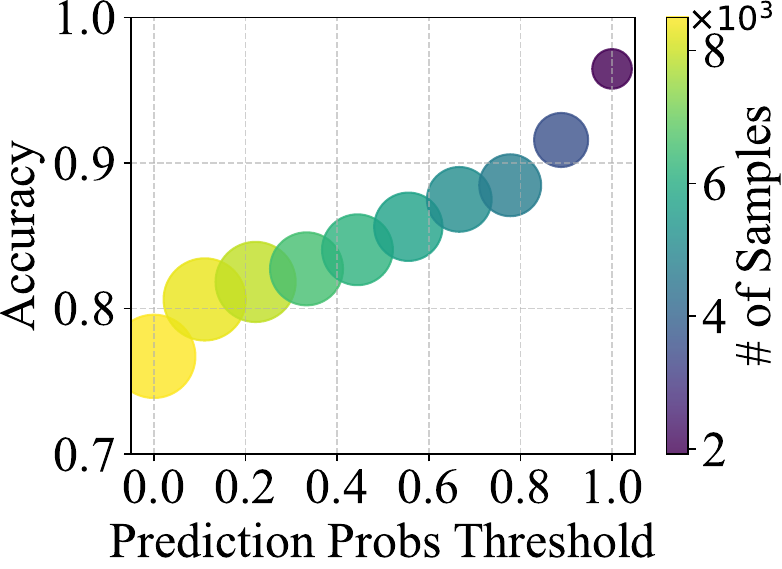}
    \caption{FHM Dataset}
    \end{subfigure}
    \hfill
    \begin{subfigure}[b]{0.491\columnwidth}
    \centering
    \includegraphics[width=\columnwidth]{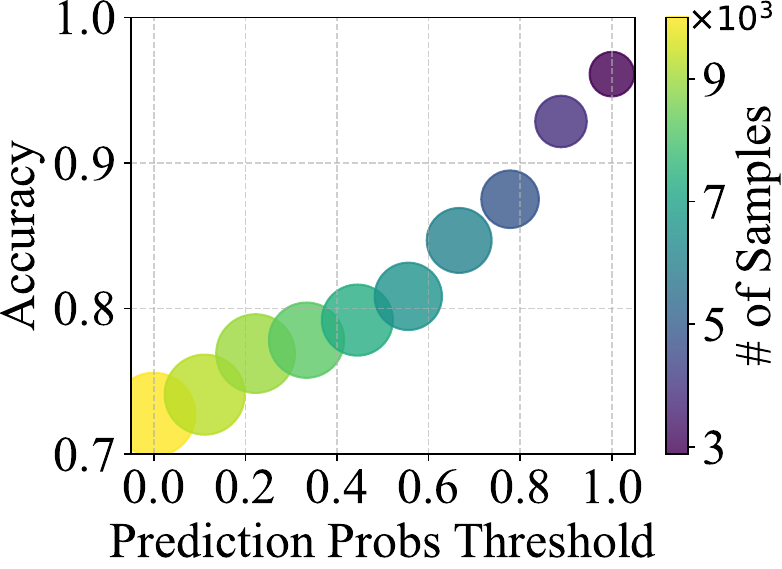}
    \caption{MAMI Dataset}
    \end{subfigure}
    \vspace{-2mm}
    \caption{The relationship between prediction probs threshold and detection accuracy on the FHM and MAMI datasets. The color and size of circles reflect the number of samples, with smaller and darker ones indicating fewer samples.}
    \label{fig:pre-exp}
    \vspace{-3mm}
\end{figure}

\subsection{Preliminary Experiment}
\label{subsec:pre-exp}
To explore the relationship between the LMMs' prediction probs (confidence) and detection accuracy in harmful meme detection, 
we conduct a preliminary experiment.
Specifically, we leverage Qwen2.5-VL-72B and follow the detection process stated in \Cref{method:pseudo_label} to perform detection on the training set of FHM and MAMI datasets.
Subsequently, we sample ten probs thresholds spanning from 0.0 to 1.0 and visualize the prediction accuracy for the meme samples whose probs exceeded each sampled value.
The size and color of each circle in the visualization represents the number of samples exceeding the corresponding probs threshold, with smaller and darker circles indicating fewer samples.
As illustrated in \Cref{fig:pre-exp},
we draw the following two observations: 
(O1) As the prediction probs threshold increases, the number of meme samples exceeding each threshold decreases, indicating that applying a probs threshold can effectively filter memes into explicit and subtle categories. 
(O2) Meme samples with higher probs values exhibit higher prediction accuracy, indicating that the LMM’s confidence is positively correlated with its ability to correctly classify harmful memes.
These observations support our hypothesis in \Cref{method:pseudo_label}: Explicit memes can be effectively isolated based on high prediction probs. 
And the accuracy of detections on these explicit memes is sufficiently high and reliable, confirming the feasibility in endowing these memes with pseudo-labels (i.e., minimal noise is introduced).

\subsection{Main Performance}
\label{subsec:main-exp}
To validate the effectiveness of our proposed \M, we compare it with \NumBaseline
competitive baselines, and the results are summarized in~\Cref{tab:main-result}. From the results, we have the following observations.

(O1) Our label-free detection framework \textbf{\M} achieves superior performance in identifying harmful memes, even slightly surpassing Label-Driven Methods across three diverse benchmarks. Remarkably, \M significantly enhances the detection capabilities of its LMM backbones (i.e., Qwen2.5-VL-72B and GPT-4o) with an average Accuracy improvement of 6.35\%.
Instead of training on the annotated memes, \M sheds light on the feasibility of utilizing the rich and powerful information from the unlabeled shallow samples for the detection of subtle and challenging ones.
To achieve competitive performance with label-driven methods, \M first accurately isolates the explicit memes and endows them with pseudo-labels.
Subsequently, an LMM agent is introduced to accumulate a wealth of detection experiences from the explicit memes and derive high-level references, which then significantly enhance the LMM's ability to detect the subtle harmful memes.
Furthermore, the merit of ``make use of local materials'' enables \M to easily and quickly respond to any newly coming memes, since it can directly extract references from part of these memes to make itself ``evolve'' to adapt to new and unseen memes.

(O2) \textbf{Label-Driven Methods} have demonstrated strong detection performance within specific datasets, primarily owing to their reliance on well-crafted supervised learning strategies on the corresponding training data.
For instance, Pro-Cap utilizes multimodal content alongside question-answering-based image captions to fine-tune models for enhanced detection accuracy.
However, they require large amount of labeled data to achieve the competitive performance, which is often impractical in real-world scenarios due to the labor-intensive nature of labeling and potential copyright constraints.
Furthermore, label-driven methods tend to specialize in capturing dataset-specific harmful patterns, which prohibits their ability to generalize effectively to emerging unseen memes and leads them fail to tackle the fast-evolving harmful memes.

(O3) While \textbf{Few-Shot Learning Methods} reduce the dependency on extensive annotated data and achieve a relatively low-resource detection mode, their performance remains suboptimal, consistently falling short of Label-Driven Methods.
The primary limitation of these approaches lies in their inability to effectively harness the rich information embedded in unlabeled memes, which significantly restricts their capability to bridge the gap between label-scarce and high detection accuracy. 
Furthermore, the detection capability of in-context few-shot learning methods are severely constrained by the limited demonstrations that can be provided to LMMs due to the limited context-length. 
This challenge is effectively mitigated by \M through its reference refinement mechanism, which distills only a set of high-level references to well represent diverse and abundant meme cases.

\begin{table}[t]
    \centering
    \caption{Ablation study on core components within \M.}
    \vspace{-3mm}
    \label{table:ablation}
        \setlength{\tabcolsep}{4pt}
    \resizebox{\linewidth}{!}{
    \begin{tabular}{lcccccc}
        \toprule
          & \multicolumn{2}{c}{\textbf{FHM}} & \multicolumn{2}{c}{\textbf{MAMI}} & \multicolumn{2}{c}{\textbf{ToxiCN}}\\
         \cmidrule(lr){2-3} \cmidrule(lr){4-5} \cmidrule(lr){6-7}
         \textbf{Variant} & \textbf{Acc} & \textbf{M-F1} & \textbf{Acc} & \textbf{M-F1} &\textbf{Acc} & \textbf{M-F1} \\
        \midrule
         w/o Confidence & 70.78 & 70.79 & 79.90 & 79.09 & 75.37 & 68.49 \\ 
         \midrule
         w/o Pairing & 72.40 & 72.33 & 78.70 & 78.69 & 74.66 & 67.13 \\
         w/o Experience & 72.00 & 71.82 & 80.00 & 79.98 & 76.99 & 70.75 \\
         w/o Reference & 69.60 & 68.46 & 78.10 & 77.90 & 73.25 & 65.84 \\
         \midrule
        \rowcolor{gray!10}
        \textbf{\M} & \textbf{75.80} & \textbf{75.79} & \textbf{81.28} & \textbf{81.25} & \textbf{79.21} & \textbf{72.51} \\
        \bottomrule
    \end{tabular}
    }
    \vspace{-5mm}
\end{table}

\subsection{Ablation Study}
\label{subsec:ablation}
We conduct a comprehensive ablation study to analyze the role of each core component in our \M, and the results are in \Cref{table:ablation}.

\subsubsection{Effect of Explicit Meme Identification}
To evaluate the effect of the \textit{Confidence-based Explicit Meme Identification mechanism}, we design a variant model \textbf{w/o Confidence}.
This variant replaces the original prediction probs (confidence) based mechanism with a simple random selection strategy, where $\tau\%$ of the memes in the training set are randomly selected as explicit ones and assigned pseudo-labels.
From the table, we observe a significant performance degradation in this variant, since the random selection of explicit memes inevitably include complex and subtle memes. 
Such memes are more prone to be given wrong pseudo-labels by the LMMs without any task-specific fine-tuning (cf. \Cref{subsec:pre-exp} for experimental evidence), thereby introducing much noise into the following processes and leading to bad performance.

\subsubsection{Effect of Agent Self-Improvement}
To validate the effect of the \textit{Pairwise Learning Guided Agent Self-Improvement paradigm}, we develop three variant models: (1) \textbf{w/o Pairing}: where the retrieval-based contrastive pairing mechanism is entirely removed, and memes from two pseudo-categories in original pairs are randomly realigned to form new pairs.
(2) \textbf{w/o Experience}: where the experience gathering process is excluded, and the references are directly derived from the initial contrastive pairs.
(3) \textbf{w/o Reference}:
where the references refinement process is eliminated, and the experiences are directly leveraged to enhance the detection capability of LMM agent.
For the \textbf{w/o Pairing} variant, the  semantically mismatched paired memes undermine the core principle of contrastive learning and degrade the quality of gathered experiences, incurring suboptimal detection performance.
For the \textbf{w/o Experience} variant, the detection references directly derived from paired memes are compromised in quality, providing LMM agent with limited guidance in detecting more complex memes.
For the \textbf{w/o Reference} variant, the gathered experiences remain case-specific and fail to offer more generalized and high-level guidance for robust detection. 
Moreover, only a few experiences can be fed to LMMs due to the limited context-length, which also limits the efficacy of this variant.

\begin{figure}[t]
    \centering
    \begin{subfigure}[b]{0.495\columnwidth}
    \centering
    \includegraphics[width=\columnwidth]{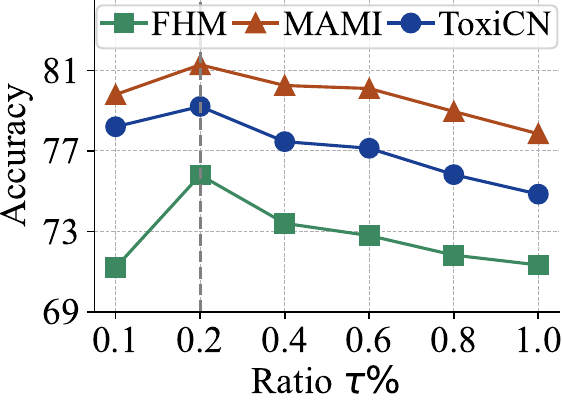}
    \vspace{-5mm}
    \caption{Hyper-Parameter $\tau$}
    \end{subfigure}
    \hfill
    \begin{subfigure}[b]{0.495\columnwidth}
    \centering
    \includegraphics[width=\columnwidth]{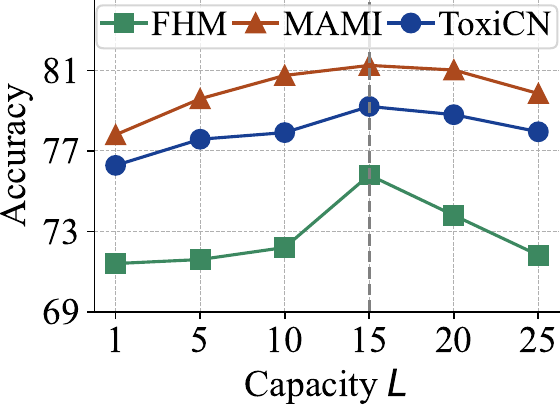}
    \vspace{-5mm}
    \caption{Hyper-Parameter $L$}
    \end{subfigure}
    \vspace{-7mm}
    \caption{Sensitivity analysis of hyper-parameters $\tau$ and $L$.}
    \label{fig:hyper-para}
    \vspace{-7mm}
\end{figure}

\subsection{Hyper-Parameter Analysis}
\label{subsec:hyper-para}
In this section, we analyze the sensitivity of two important hyper-parameters within \M: the confidence selection ratio $\tau$ and reference set capacity $L$.
From the results presented in \Cref{fig:hyper-para}, we draw the following conclusions for these two hyper-parameters.

(C1) The ratio $\tau$ determines the proportion of samples with the highest prediction confidence that are selected as explicit memes. 
As shown in \Cref{fig:hyper-para}(a), an excessively low value of $\tau$ results in suboptimal performance. 
This is because only a very small subset of samples is selected as explicit memes, thereby providing insufficient detection references. 
Conversely, an extreme large value of $\tau$ also degrades the performance due to the inclusion of numerous hard and complex samples that are incorrectly identified as explicit memes. 
Consequently, we adopt $\tau = 0.2$ across all three datasets.

(C2) The capacity $L$ represents the maximum number of references that can be maintained and utilized by the LMM agent. 
As presented in \Cref{fig:hyper-para}(b), the detection performance initially improves with an increase in $L$, as a larger capacity provides the agent with more informative and diverse references. 
However, excessively large reference capacities can degrade performance, as the LMM agent may incorporate redundant or even noisy references, compromising the overall quality of the reference set. 
Based on these observations,
we set $L = 15$ on all three datasets.

\begin{figure}[t]
    \centering
    \includegraphics[width=\columnwidth]{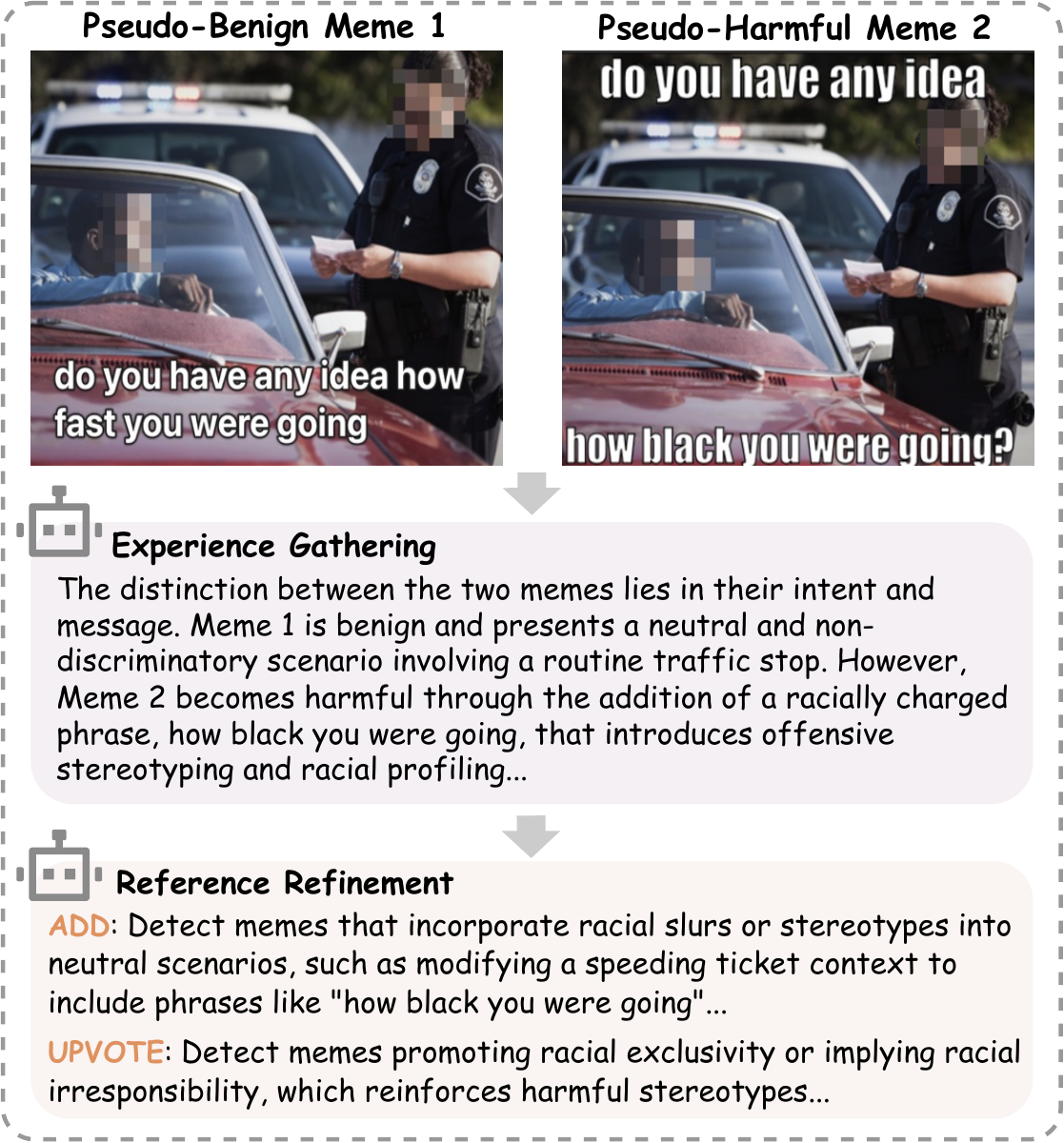}
    \vspace{-6mm}
    \caption{Case study on experience gathering and reference refinement: A contrastive pair is analyzed, including both pseudo-benign and pseudo-harmful memes, alongside the experiences and references derived from this pair.
    }
    \label{fig:case-1}
    \vspace{-2mm}
\end{figure}

\subsection{Qualitative Analysis}
\label{subsec:qualitative}
In this section, to further delve into how \M positions an LMM agent to gather case-based experiences from explicit memes, refine high-level detection references, and leverage these references to tackle more challenging memes, we conduct a comprehensive qualitative analysis in the following subsections. 

\subsubsection{Case Study on Experience Gathering \& Reference Refinement}
First, we perform a case study to qualitatively analyze the effectiveness of the experience gathering and reference refinement process.
Specifically, we randomly select a contrastive pair from the FHM dataset, which include two explicit memes with contradict pseudo-labels.
As shown in \Cref{fig:case-1}, the LMM agent analyzes and discerns the nuanced differences between the two contrastive memes: The first meme describes a neutral scenario of traffic stop, delivering no malicious message, while the second meme—despite sharing the same neutral visual content—conveys racial discrimination speech by slightly modifying the captions.
Building on these experiences, the agent refines a new detection reference that specifically targets scenarios where racial slurs are introduced by altering neutral contexts. 
Additionally, the agent increases the importance of an existing reference, which delivers detection criteria in the similar scenario (racial discrimination) but from different angle.

\begin{figure}[t]
    \centering
    \includegraphics[width=\columnwidth]{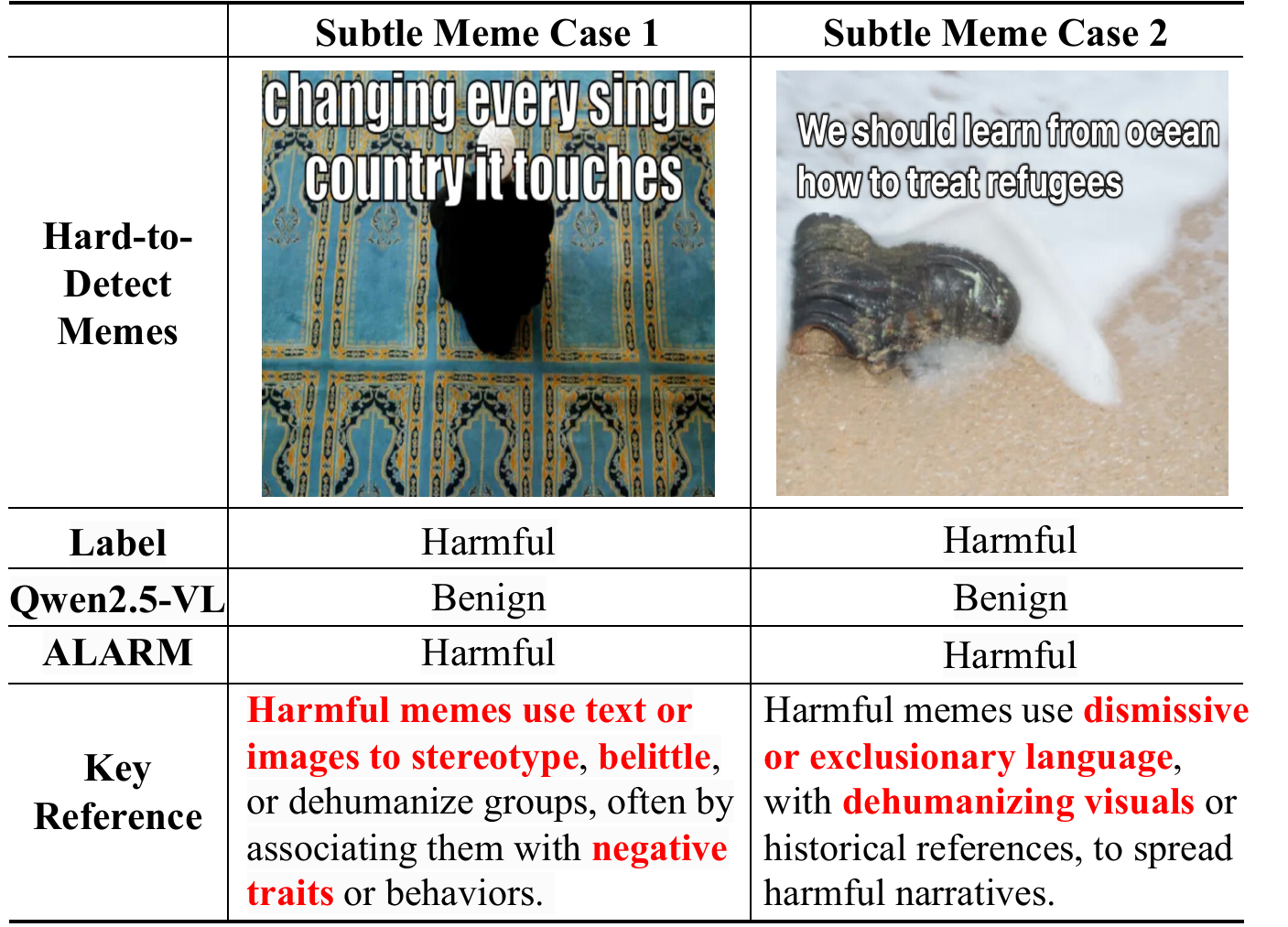}
    \vspace{-6mm}
    \caption{Case study on LMM agent self-improvement: The Qwen2.5-VL-72B model misclassifies both subtle memes as benign, while \M successfully identifies the harmful content by equipping the self-generated detection references.
    }
    \label{fig:case-2}
    \vspace{-2mm}
\end{figure}

\subsubsection{Case Study on LMM Agent Self-Improvement}
Second, we conduct another case study to analyze how \M enhances the LMM in improving the detection capability on hard meme samples through the guidance of self-generated references.
Specifically, we randomly select two subtle and hard-to-detect harmful memes from the FHM dataset, along with the detection results of both vanilla LMM backbone --- Qwen2.5-VL-72B and \M.
As illustrated in \Cref{fig:case-2}, the first subtle meme challenges vanilla LMMs because the benign-looking silhouette offers no explicit visual cue linking to the text's implication that a particular group ``changes every country it touches.'' Lacking this connection, the model fails to detect the latent stereotype. 
By introducing the reference ``associating groups with negative traits or behaviors,'' \M guides the LMM to attend to such implicit stereotyping, enabling correct harmfulness detection.
The second subtle meme similarly employs an innocuous coastal scene that may soften the impact of the exclusionary message suggesting refugees should be treated like objects washed away by the ocean. 
With the reference highlighting ``dismissive or dehumanizing visuals used to advance harmful narratives,'' \M enables the LMM to scrutinize such seemingly harmless imagery, allowing it to accurately classify the meme as harmful.

\begin{figure}[t]
    \centering
    \begin{subfigure}[b]{0.492\columnwidth}
    \centering
    \includegraphics[width=\columnwidth]{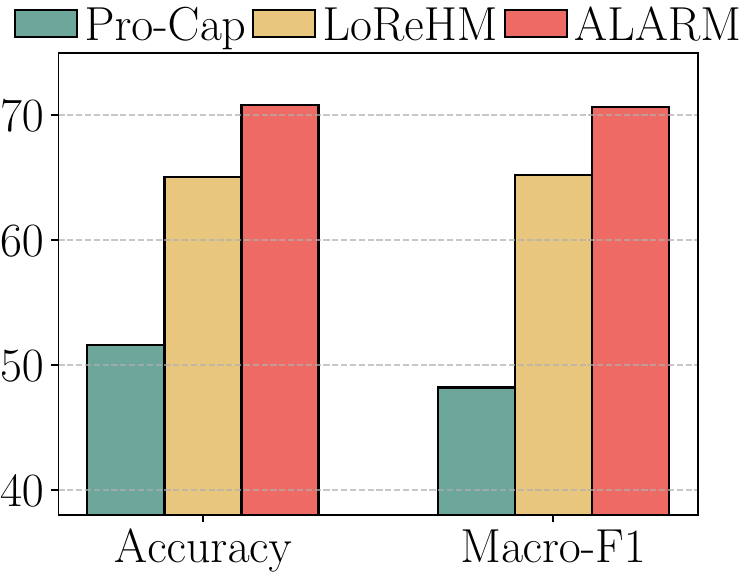}
    \caption{Testing on FHM Dataset}
    \end{subfigure}
    \hfill
    \begin{subfigure}[b]{0.492\columnwidth}
    \centering
    \includegraphics[width=\columnwidth]{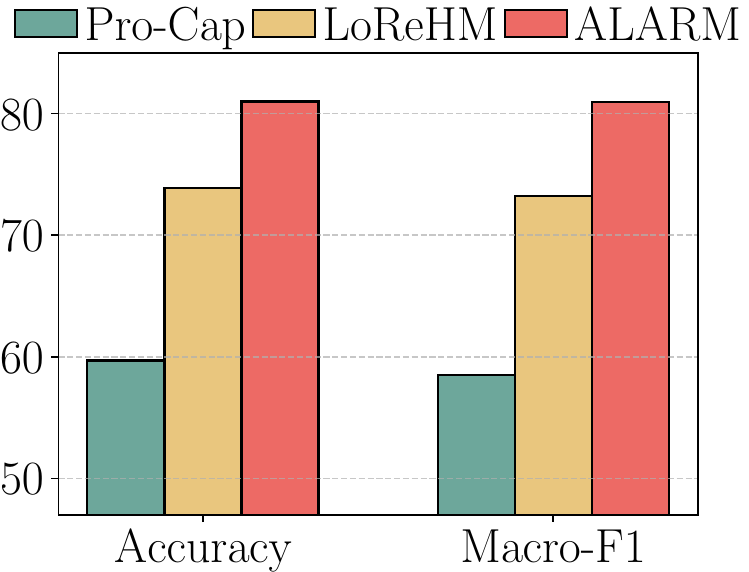}
    \caption{Testing on MAMI Dataset}
    \end{subfigure}
    \vspace{-7mm}
    \caption{Generalizability comparison between \M and two competitive baselines, Pro-Cap and LoReHM.}
    \label{fig:general}
    \vspace{-3mm}
\end{figure}

\subsection{Generalization on Unseen Memes}
\label{subsec:general}
To evaluate the generalizability of \M in addressing newly emerging and unseen memes, we conduct an experiment and compare its performance with two competitive baselines: Pro-Cap and LoReHM\footnote{Both \M and LoReHM employ Qwen2.5-VL-72B as LMM for fair comparison.}. 
Specifically, we use test set of datasets FHM and MAMI as unseen data, respectively. 
For label-driven method Pro-Cap, we provide labeled data from a different dataset (i.e., testing on the FHM dataset uses data sourced from training set of MAMI, vice versa). 
Similarly, for few-shot method LoReHM, we provide demonstrations from the other dataset.
In contrast, \M operates without any labeled data or demonstrations, instead leveraging unannotated test data to enhance detection. 
As shown in \Cref{fig:general}, \M achieves the best generalization performance by acquiring valuable detection references directly from the test memes and equipping these references to realize ``self-improvement'', showcasing a quick-responding solution to the newly coming harmful memes.

\subsection{Transferability to Smaller LMMs}
\label{subsec:transfer}
\begin{figure}[t]
    \centering
    \begin{subfigure}[b]{0.488\columnwidth}
    \centering
    \includegraphics[width=\columnwidth]{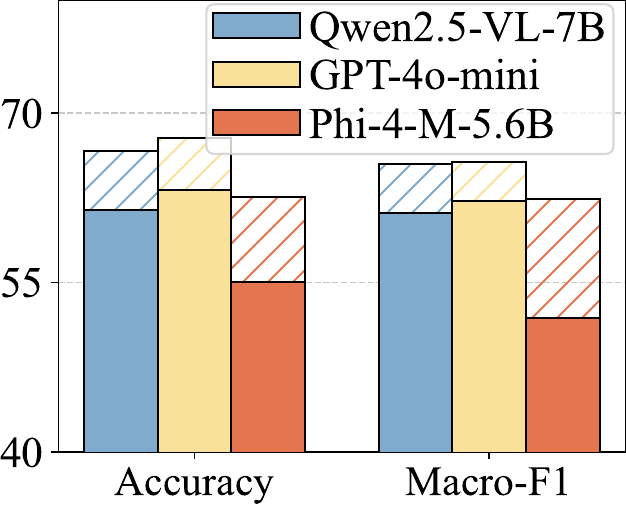}
    \vspace{-5mm}
    \caption{FHM Dataset}
    \end{subfigure}
    \hfill
    \begin{subfigure}[b]{0.488\columnwidth}
    \centering
    \includegraphics[width=\columnwidth]{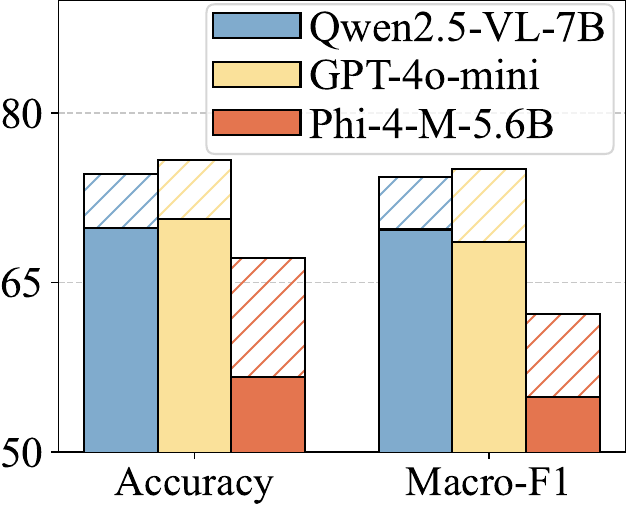}
    \vspace{-5mm}
    \caption{MAMI Dataset}
    \end{subfigure}
    \vspace{-3mm}
    \caption{Transferability analysis of the detection references generated from \M to three smaller LMMs. The hatch area above the original performance represents the improvement after equipping the references.}
    \label{fig:transfer}
    \vspace{-3mm}
\end{figure}

To assess the quality and scalability of the detection references autonomously generated by the LMM agent in \M, we conduct a down-scaling transfer experiment on the FHM and MAMI datasets.
In this experiment, detection references generated by \M using the large LMM backbone Qwen2.5-VL-72B are transferred to three smaller and popular LMMs: Qwen2.5-VL-7B~\cite{bai2025qwen2}, GPT-4o-mini~\cite{openai2024gpt}, and Phi-4-Multimodal-5.6B~\cite{abouelenin2025phi}. 
We then evaluate the performance variability across these smaller models.
As illustrated in \Cref{fig:transfer}, all three smaller LMMs show significant performance improvements when leveraging the references, showcasing the high quality and strong scalability of the detection references.
These improvements highlight the potential of \M in effectively transferring detection knowledge to smaller and resource-constrained models, making scalable lightweight harmful meme detection a reality.

\subsection{Efficiency Analysis}
\label{app-eff}
\begin{figure}[t]
    \centering
    \includegraphics[width=0.99\columnwidth]{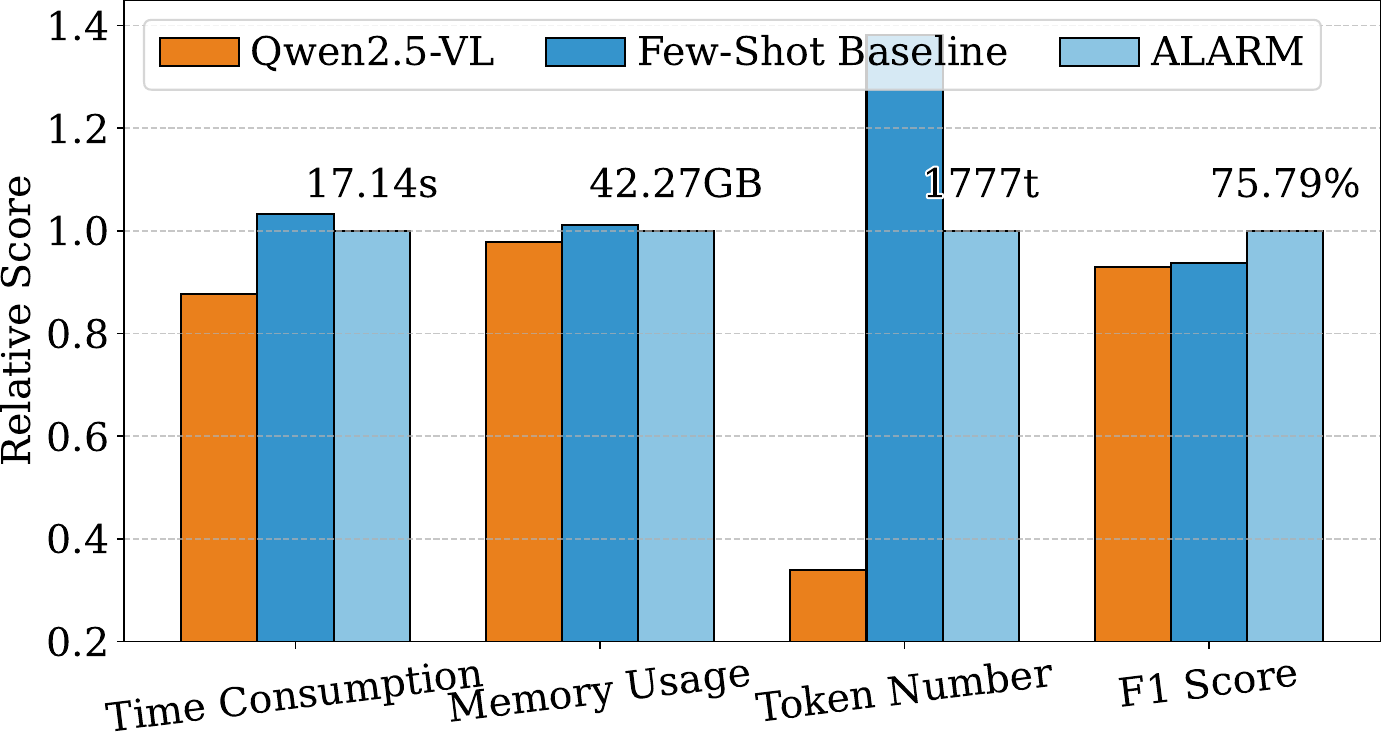}
    \vspace{-3mm}
    \caption{Efficiency comparison among \M, the vanilla Qwen2.5-VL-72B, and a few-shot learning baseline.
    }
    \label{fig:efficiency}
    \vspace{-4mm}
\end{figure}

Given the label- and gradient-free characteristics of our \M, which is founded on LMMs, we conduct an empirical analysis to assess its efficiency during inference.
Specifically, Qwen2.5-VL-72B~\cite{bai2025qwen2} is adopted as our LMM backbone and the efficiency assessment focuses on three aspects, including the inference time, GPU memory usage, and the number of input tokens required.
We compare its efficiency against the vanilla LMM Qwen2.5-VL-72B model and a few-shot learning baseline implemented on the Qwen2.5-VL-72B.
For vanilla Qwen2.5-VL-72B, we provide it with a simple CoT-based prompt to analyze the multimodal content in the memes and detect whether there are harmful content exist.
In contrast, the few-shot learning baseline is fed with 50 textual in-context examples to guide the detection.
We visualize the inference time consumption, GPU memory usage, input token count, and performance on the FHM dataset for each model in \Cref{fig:efficiency}. 
Each metric is normalized to a relative scale using our \M as the reference baseline. Additionally, the actual values of each metric for \M are annotated in the figure.
From the figure, we observe that the vanilla Qwen2.5-VL-72B achieves the lowest computational overhead due to its concise prompt design, which minimizes the number of input tokens during inference.
However, this efficiency leads to significantly compromised performance, indicating its inability to detect complex and challenging memes.
On the other hand, the few-shot learning baseline demonstrates a marked improvement in detection performance over the vanilla Qwen2.5-VL-72B. 
This gain, however, comes with substantial computational overhead, as the inclusion of 50 in-context examples largely increases the count of input tokens processed by the Qwen2.5-VL-72B.
Our framework \M largely improves the performance of the Qwen2.5-VL-72B by equipping it with a set of valuable detection references, realizing a remarkable balance between efficiency and performance.

\section{Conclusion}
\label{sec:conclusion}

In this study, to address the reliance on massive labeled data in harmful meme detection, we propose the framework \M.
It demonstrates the feasibility of label-free detection by utilizing informative knowledge from unlabeled explicit samples to tackle more challenging ones.
\M first introduces a \textit{Confidence-based Explicit Meme Identification mechanism}, which realizes the extraction and pseudo-labeling of explicit memes.
Based on this, a \textit{Pairwise Learning Guided Agent Self-Improvement paradigm} is proposed, which realigns these explicit memes into contrastive pairs and facilitates an LMM agent to derive valuable references from these pairs.
Finally, the self-generated references empower the LMM agent to achieve ``self-improvement'', enabling it to detect more challenging memes.
Extensive experiments on three datasets highlight the efficacy of \M, offering a novel path toward label-free detection.

\section*{Acknowledgments}
This work was supported by National Natural Science Foundation of China (Grant No. 62572097, No. 62176043, and No. U22A2097).

\newpage
  
\bibliographystyle{ACM-Reference-Format}
\balance
\bibliography{main}


\begin{thebibliography}{44}

%%% the Web address.

\ifx \showCODEN    \undefined \def \showCODEN     #1{\unskip}     \fi
\ifx \showISBNx    \undefined \def \showISBNx     #1{\unskip}     \fi
\ifx \showISBNxiii \undefined \def \showISBNxiii  #1{\unskip}     \fi
\ifx \showISSN     \undefined \def \showISSN      #1{\unskip}     \fi
\ifx \showLCCN     \undefined \def \showLCCN      #1{\unskip}     \fi
\ifx \shownote     \undefined \def \shownote      #1{#1}          \fi
\ifx \showarticletitle \undefined \def \showarticletitle #1{#1}   \fi
\ifx \showURL      \undefined \def \showURL       {\relax}        \fi
% invisible to TeX
\providecommand\bibfield[2]{#2}
\providecommand\bibinfo[2]{#2}
\providecommand\natexlab[1]{#1}
\providecommand\showeprint[2][]{arXiv:#2}

\bibitem[Abouelenin et~al\mbox{.}(2025)]%
        {abouelenin2025phi}
\bibfield{author}{\bibinfo{person}{Abdelrahman Abouelenin}, \bibinfo{person}{Atabak Ashfaq}, \bibinfo{person}{Adam Atkinson}, \bibinfo{person}{Hany Awadalla}, \bibinfo{person}{Nguyen Bach}, \bibinfo{person}{Jianmin Bao}, \bibinfo{person}{Alon Benhaim}, \bibinfo{person}{Martin Cai}, \bibinfo{person}{Vishrav Chaudhary}, {et~al\mbox{.}}} \bibinfo{year}{2025}\natexlab{}.
\newblock \showarticletitle{Phi-4-mini technical report: Compact yet powerful multimodal language models via mixture-of-loras}.
\newblock \bibinfo{journal}{\emph{arXiv preprint arXiv:2503.01743}} (\bibinfo{year}{2025}).
\newblock


\bibitem[Abramson et~al\mbox{.}(2022)]%
        {abramson2022improving}
\bibfield{author}{\bibinfo{person}{Josh Abramson}, \bibinfo{person}{Arun Ahuja}, \bibinfo{person}{Federico Carnevale}, \bibinfo{person}{Petko Georgiev}, \bibinfo{person}{Alex Goldin}, \bibinfo{person}{Alden Hung}, \bibinfo{person}{Jessica Landon}, \bibinfo{person}{Jirka Lhotka}, \bibinfo{person}{Timothy Lillicrap}, \bibinfo{person}{Alistair Muldal}, {et~al\mbox{.}}} \bibinfo{year}{2022}\natexlab{}.
\newblock \showarticletitle{Improving multimodal interactive agents with reinforcement learning from human feedback}.
\newblock \bibinfo{journal}{\emph{arXiv preprint arXiv:2211.11602}} (\bibinfo{year}{2022}).
\newblock


\bibitem[Achiam et~al\mbox{.}(2023)]%
        {achiam2023gpt}
\bibfield{author}{\bibinfo{person}{Josh Achiam}, \bibinfo{person}{Steven Adler}, \bibinfo{person}{Sandhini Agarwal}, \bibinfo{person}{Lama Ahmad}, \bibinfo{person}{Ilge Akkaya}, \bibinfo{person}{Florencia~Leoni Aleman}, \bibinfo{person}{Diogo Almeida}, \bibinfo{person}{Janko Altenschmidt}, \bibinfo{person}{Sam Altman}, {et~al\mbox{.}}} \bibinfo{year}{2023}\natexlab{}.
\newblock \showarticletitle{Gpt-4 technical report}.
\newblock \bibinfo{journal}{\emph{arXiv preprint arXiv:2303.08774}} (\bibinfo{year}{2023}).
\newblock


\bibitem[Alayrac et~al\mbox{.}(2022)]%
        {alayrac2022flamingo}
\bibfield{author}{\bibinfo{person}{Jean-Baptiste Alayrac}, \bibinfo{person}{Jeff Donahue}, \bibinfo{person}{Pauline Luc}, \bibinfo{person}{Antoine Miech}, \bibinfo{person}{Iain Barr}, \bibinfo{person}{Yana Hasson}, \bibinfo{person}{Karel Lenc}, \bibinfo{person}{A. Mensch}, \bibinfo{person}{Katie Millican}, \bibinfo{person}{Malcolm Reynolds}, \bibinfo{person}{Roman Ring}, \bibinfo{person}{Eliza Rutherford}, \bibinfo{person}{Serkan Cabi}, \bibinfo{person}{Tengda Han}, \bibinfo{person}{Zhitao Gong}, \bibinfo{person}{Sina Samangooei}, \bibinfo{person}{Marianne Monteiro}, \bibinfo{person}{Jacob Menick}, \bibinfo{person}{Sebastian Borgeaud}, \bibinfo{person}{Andy Brock}, \bibinfo{person}{Aida Nematzadeh}, \bibinfo{person}{Sahand Sharifzadeh}, \bibinfo{person}{Mikolaj Binkowski}, \bibinfo{person}{Ricardo Barreira}, \bibinfo{person}{O. Vinyals}, \bibinfo{person}{Andrew Zisserman}, {and} \bibinfo{person}{K. Simonyan}.} \bibinfo{year}{2022}\natexlab{}.
\newblock \showarticletitle{Flamingo: a {Visual} {Language} {Model} for {Few}-{Shot} {Learning}}. In \bibinfo{booktitle}{\emph{Conference on {Neural} {Information} {Processing} {Systems} ({NeurIPS})}}, Vol.~\bibinfo{volume}{abs/2204.14198}.
\newblock


\bibitem[Bai et~al\mbox{.}(2025)]%
        {bai2025qwen2}
\bibfield{author}{\bibinfo{person}{Shuai Bai}, \bibinfo{person}{Keqin Chen}, \bibinfo{person}{Xuejing Liu}, \bibinfo{person}{Jialin Wang}, \bibinfo{person}{Wenbin Ge}, \bibinfo{person}{Sibo Song}, \bibinfo{person}{Kai Dang}, \bibinfo{person}{Peng Wang}, \bibinfo{person}{Shijie Wang}, \bibinfo{person}{Jun Tang}, {et~al\mbox{.}}} \bibinfo{year}{2025}\natexlab{}.
\newblock \showarticletitle{Qwen2. 5-vl technical report}.
\newblock \bibinfo{journal}{\emph{arXiv preprint arXiv:2502.13923}} (\bibinfo{year}{2025}).
\newblock


\bibitem[Cao et~al\mbox{.}(2023)]%
        {cao2023procap}
\bibfield{author}{\bibinfo{person}{Rui Cao}, \bibinfo{person}{Ming~Shan Hee}, \bibinfo{person}{Adriel Kuek}, \bibinfo{person}{Wen-Haw Chong}, \bibinfo{person}{Roy Ka-Wei Lee}, {and} \bibinfo{person}{Jing Jiang}.} \bibinfo{year}{2023}\natexlab{}.
\newblock \showarticletitle{Pro-{Cap}: Leveraging a {Frozen} {Vision}-{Language} {Model} for {Hateful} {Meme} {Detection}}. In \bibinfo{booktitle}{\emph{ACM {International} {Conference} on {Multimedia} ({MM})}}. ACM.
\newblock


\bibitem[Cao et~al\mbox{.}(2022)]%
        {cao2022prompting}
\bibfield{author}{\bibinfo{person}{Rui Cao}, \bibinfo{person}{Roy Ka-Wei Lee}, \bibinfo{person}{Wen-Haw Chong}, {and} \bibinfo{person}{Jing Jiang}.} \bibinfo{year}{2022}\natexlab{}.
\newblock \showarticletitle{Prompting for {Multimodal} {Hateful} {Meme} {Classification}.}. In \bibinfo{booktitle}{\emph{Conference on {Empirical} {Methods} in {Natural} {Language} {Processing} ({EMNLP})}}. \bibinfo{pages}{321--332}.
\newblock


\bibitem[Cao et~al\mbox{.}(2024)]%
        {cao2024modularized}
\bibfield{author}{\bibinfo{person}{Rui Cao}, \bibinfo{person}{Roy Ka-Wei Lee}, {and} \bibinfo{person}{Jing Jiang}.} \bibinfo{year}{2024}\natexlab{}.
\newblock \showarticletitle{Modularized {Networks} for {Few}-shot {Hateful} {Meme} {Detection}}. In \bibinfo{booktitle}{\emph{Proceedings of the {ACM} {Web} {Conference}}}. ACM.
\newblock


\bibitem[Dorigo and Colombetti(1994)]%
        {dorigo1994robot}
\bibfield{author}{\bibinfo{person}{Marco Dorigo} {and} \bibinfo{person}{Marco Colombetti}.} \bibinfo{year}{1994}\natexlab{}.
\newblock \showarticletitle{Robot shaping: Developing autonomous agents through learning}.
\newblock \bibinfo{journal}{\emph{Artificial intelligence}} \bibinfo{volume}{71}, \bibinfo{number}{2} (\bibinfo{year}{1994}), \bibinfo{pages}{321--370}.
\newblock


\bibitem[Fang et~al\mbox{.}(2024)]%
        {fang2024eva}
\bibfield{author}{\bibinfo{person}{Yuxin Fang}, \bibinfo{person}{Quan Sun}, \bibinfo{person}{Xinggang Wang}, \bibinfo{person}{Tiejun Huang}, \bibinfo{person}{Xinlong Wang}, {and} \bibinfo{person}{Yue Cao}.} \bibinfo{year}{2024}\natexlab{}.
\newblock \showarticletitle{Eva-02: A visual representation for neon genesis}.
\newblock \bibinfo{journal}{\emph{Image and Vision Computing}}  \bibinfo{volume}{149} (\bibinfo{year}{2024}), \bibinfo{pages}{105171}.
\newblock


\bibitem[Fersini et~al\mbox{.}(2022)]%
        {fersini2022semeval}
\bibfield{author}{\bibinfo{person}{Elisabetta Fersini}, \bibinfo{person}{Francesca Gasparini}, \bibinfo{person}{Giulia Rizzi}, \bibinfo{person}{Aurora Saibene}, \bibinfo{person}{Berta Chulvi}, \bibinfo{person}{Paolo Rosso}, \bibinfo{person}{Alyssa Lees}, {and} \bibinfo{person}{Jeffrey Sorensen}.} \bibinfo{year}{2022}\natexlab{}.
\newblock \showarticletitle{SemEval-2022 {Task} 5: Multimedia {Automatic} {Misogyny} {Identification}}. In \bibinfo{booktitle}{\emph{Proceedings of the 16th {International} {Workshop} on {Semantic} {Evaluation} ({SemEval}-2022)}}. Association for Computational Linguistics.
\newblock


\bibitem[Han et~al\mbox{.}(2024)]%
        {han2024multimodal}
\bibfield{author}{\bibinfo{person}{Soyeon~Caren Han}, \bibinfo{person}{Feiqi Cao}, \bibinfo{person}{Josiah Poon}, {and} \bibinfo{person}{Roberto Navigli}.} \bibinfo{year}{2024}\natexlab{}.
\newblock \showarticletitle{Multimodal Large Language Models and Tunings: Vision, Language, Sensors, Audio, and Beyond}. In \bibinfo{booktitle}{\emph{ACM {International} {Conference} on {Multimedia} ({MM})}}. \bibinfo{pages}{11294--11295}.
\newblock


\bibitem[Hee et~al\mbox{.}(2024)]%
        {hee2024bridging}
\bibfield{author}{\bibinfo{person}{Ming~Shan Hee}, \bibinfo{person}{Aditi Kumaresan}, {and} \bibinfo{person}{Roy Ka-Wei Lee}.} \bibinfo{year}{2024}\natexlab{}.
\newblock \showarticletitle{Bridging {Modalities}: Enhancing {Cross}-{Modality} {Hate} {Speech} {Detection} with {Few}-{Shot} {In}-{Context} {Learning}}. In \bibinfo{booktitle}{\emph{Conference on {Empirical} {Methods} in {Natural} {Language} {Processing} ({EMNLP})}}. \bibinfo{pages}{7785--7799}.
\newblock


\bibitem[Huang et~al\mbox{.}(2023)]%
        {huang2023large}
\bibfield{author}{\bibinfo{person}{Jiaxin Huang}, \bibinfo{person}{Shixiang Gu}, \bibinfo{person}{Le Hou}, \bibinfo{person}{Yuexin Wu}, \bibinfo{person}{Xuezhi Wang}, \bibinfo{person}{Hongkun Yu}, {and} \bibinfo{person}{Jiawei Han}.} \bibinfo{year}{2023}\natexlab{}.
\newblock \showarticletitle{Large Language Models Can Self-Improve}. In \bibinfo{booktitle}{\emph{Conference on {Empirical} {Methods} in {Natural} {Language} {Processing} ({EMNLP})}}. \bibinfo{pages}{1051--1068}.
\newblock


\bibitem[Huang et~al\mbox{.}(2024)]%
        {huang2024towards}
\bibfield{author}{\bibinfo{person}{Jianzhao Huang}, \bibinfo{person}{Hongzhan Lin}, \bibinfo{person}{Liu Ziyan}, \bibinfo{person}{Ziyang Luo}, \bibinfo{person}{Guang Chen}, {and} \bibinfo{person}{Jing Ma}.} \bibinfo{year}{2024}\natexlab{}.
\newblock \showarticletitle{Towards {Low}-{Resource} {Harmful} {Meme} {Detection} with {LMM} {Agents}}. In \bibinfo{booktitle}{\emph{Conference on {Empirical} {Methods} in {Natural} {Language} {Processing} ({EMNLP})}}.
\newblock


\bibitem[Hurst et~al\mbox{.}(2024)]%
        {openai2024gpt}
\bibfield{author}{\bibinfo{person}{Aaron Hurst}, \bibinfo{person}{Adam Lerer}, \bibinfo{person}{Adam~P Goucher}, \bibinfo{person}{Adam Perelman}, \bibinfo{person}{Aditya Ramesh}, \bibinfo{person}{Aidan Clark}, \bibinfo{person}{AJ Ostrow}, \bibinfo{person}{Akila Welihinda}, \bibinfo{person}{Alan Hayes}, \bibinfo{person}{Alec Radford}, {et~al\mbox{.}}} \bibinfo{year}{2024}\natexlab{}.
\newblock \showarticletitle{Gpt-4o system card}.
\newblock \bibinfo{journal}{\emph{arXiv preprint arXiv:2410.21276}} (\bibinfo{year}{2024}).
\newblock


\bibitem[Ji et~al\mbox{.}(2023)]%
        {ji2023identifying}
\bibfield{author}{\bibinfo{person}{Junhui Ji}, \bibinfo{person}{Wei Ren}, {and} \bibinfo{person}{Usman Naseem}.} \bibinfo{year}{2023}\natexlab{}.
\newblock \showarticletitle{Identifying creative harmful memes via prompt based approach}. In \bibinfo{booktitle}{\emph{Proceedings of the {ACM} {Web} {Conference}}}.
\newblock


\bibitem[Kiela et~al\mbox{.}(2020)]%
        {kiela2020hateful}
\bibfield{author}{\bibinfo{person}{Douwe Kiela}, \bibinfo{person}{Hamed Firooz}, \bibinfo{person}{Aravind Mohan}, \bibinfo{person}{Vedanuj Goswami}, \bibinfo{person}{Amanpreet Singh}, \bibinfo{person}{Pratik Ringshia}, {and} \bibinfo{person}{Davide Testuggine}.} \bibinfo{year}{2020}\natexlab{}.
\newblock \showarticletitle{The {Hateful} {Memes} {Challenge}: Detecting {Hate} {Speech} in {Multimodal} {Memes}}. In \bibinfo{booktitle}{\emph{Conference on {Neural} {Information} {Processing} {Systems} ({NeurIPS})}}, Vol.~\bibinfo{volume}{abs/2005.04790}.
\newblock


\bibitem[Lang et~al\mbox{.}(2025)]%
        {lang2025biting}
\bibfield{author}{\bibinfo{person}{Jian Lang}, \bibinfo{person}{Rongpei Hong}, \bibinfo{person}{Jin Xu}, \bibinfo{person}{Yili Li}, \bibinfo{person}{Xovee Xu}, {and} \bibinfo{person}{Fan Zhou}.} \bibinfo{year}{2025}\natexlab{}.
\newblock \showarticletitle{Biting Off More Than You Can Detect: Retrieval-Augmented Multimodal Experts for Short Video Hate Detection}. In \bibinfo{booktitle}{\emph{The {Web} {Conference} ({WWW})}}. ACM.
\newblock


\bibitem[Langley et~al\mbox{.}(2017)]%
        {langley2017explainable}
\bibfield{author}{\bibinfo{person}{Pat Langley}, \bibinfo{person}{Ben Meadows}, \bibinfo{person}{Mohan Sridharan}, {and} \bibinfo{person}{Dongkyu Choi}.} \bibinfo{year}{2017}\natexlab{}.
\newblock \showarticletitle{Explainable agency for intelligent autonomous systems}. In \bibinfo{booktitle}{\emph{Proceedings of the AAAI Conference on Artificial Intelligence (AAAI)}}, Vol.~\bibinfo{volume}{31}. \bibinfo{pages}{4762--4763}.
\newblock


\bibitem[Lee et~al\mbox{.}(2021)]%
        {lee2021disentangling}
\bibfield{author}{\bibinfo{person}{Roy Ka-Wei Lee}, \bibinfo{person}{Rui Cao}, \bibinfo{person}{Ziqing Fan}, \bibinfo{person}{Jing Jiang}, {and} \bibinfo{person}{Wen-Haw Chong}.} \bibinfo{year}{2021}\natexlab{}.
\newblock \showarticletitle{Disentangling hate in online memes}. In \bibinfo{booktitle}{\emph{ACM {International} {Conference} on {Multimedia} ({MM})}}. \bibinfo{pages}{5138--5147}.
\newblock


\bibitem[Liang et~al\mbox{.}(2020)]%
        {liang2020we}
\bibfield{author}{\bibinfo{person}{Jian Liang}, \bibinfo{person}{Dapeng Hu}, {and} \bibinfo{person}{Jiashi Feng}.} \bibinfo{year}{2020}\natexlab{}.
\newblock \showarticletitle{Do we really need to access the source data? source hypothesis transfer for unsupervised domain adaptation}. In \bibinfo{booktitle}{\emph{International Conference on Machine Learning (ICML)}}. PMLR, \bibinfo{pages}{6028--6039}.
\newblock


\bibitem[Lin et~al\mbox{.}(2024)]%
        {lin2024towards}
\bibfield{author}{\bibinfo{person}{Hongzhan Lin}, \bibinfo{person}{Ziyang Luo}, \bibinfo{person}{Wei Gao}, \bibinfo{person}{Jing Ma}, \bibinfo{person}{Bo Wang}, {and} \bibinfo{person}{Ruichao Yang}.} \bibinfo{year}{2024}\natexlab{}.
\newblock \showarticletitle{Towards {Explainable} {Harmful} {Meme} {Detection} through {Multimodal} {Debate} between {Large} {Language} {Models}}. In \bibinfo{booktitle}{\emph{Proceedings of the {ACM} {Web} {Conference}}}. ACM, \bibinfo{pages}{2359--2370}.
\newblock


\bibitem[Lin et~al\mbox{.}(2023)]%
        {lin2023beneath}
\bibfield{author}{\bibinfo{person}{Hongzhan Lin}, \bibinfo{person}{Ziyang Luo}, \bibinfo{person}{Jing Ma}, {and} \bibinfo{person}{Long Chen}.} \bibinfo{year}{2023}\natexlab{}.
\newblock \showarticletitle{Beneath the {Surface}: Unveiling {Harmful} {Memes} with {Multimodal} {Reasoning} {Distilled} from {Large} {Language} {Models}.}. In \bibinfo{booktitle}{\emph{Conference on {Empirical} {Methods} in {Natural} {Language} {Processing} ({EMNLP})}}. \bibinfo{pages}{9114--9128}.
\newblock


\bibitem[Liu et~al\mbox{.}(2025)]%
        {liu2025visualagentbench}
\bibfield{author}{\bibinfo{person}{Xiao Liu}, \bibinfo{person}{Tianjie Zhang}, \bibinfo{person}{Yu Gu}, \bibinfo{person}{Iat~Long Iong}, \bibinfo{person}{Yifan Xu}, \bibinfo{person}{Xixuan Song}, \bibinfo{person}{Shudan Zhang}, \bibinfo{person}{Hanyu Lai}, \bibinfo{person}{Xinyi Liu}, \bibinfo{person}{Hanlin Zhao}, {et~al\mbox{.}}} \bibinfo{year}{2025}\natexlab{}.
\newblock \showarticletitle{Visualagentbench: Towards large multimodal models as visual foundation agents}. In \bibinfo{booktitle}{\emph{International {Conference} on {Learning} {Representations} ({ICLR})}}.
\newblock


\bibitem[Lu et~al\mbox{.}(2024)]%
        {lu2024towards}
\bibfield{author}{\bibinfo{person}{Junyu Lu}, \bibinfo{person}{Bo Xu}, \bibinfo{person}{Xiaokun Zhang}, \bibinfo{person}{Hongbo Wang}, \bibinfo{person}{Haohao Zhu}, \bibinfo{person}{Dongyu Zhang}, \bibinfo{person}{Liang Yang}, {and} \bibinfo{person}{Hongfei Lin}.} \bibinfo{year}{2024}\natexlab{}.
\newblock \showarticletitle{Towards {Comprehensive} {Detection} of {Chinese} {Harmful} {Memes}.}. In \bibinfo{booktitle}{\emph{Conference on {Neural} {Information} {Processing} {Systems} ({NeurIPS})}}.
\newblock


\bibitem[Madaan et~al\mbox{.}(2023)]%
        {madaan2023selfrefine}
\bibfield{author}{\bibinfo{person}{Aman Madaan}, \bibinfo{person}{Niket Tandon}, \bibinfo{person}{Prakhar Gupta}, \bibinfo{person}{Skyler Hallinan}, \bibinfo{person}{Luyu Gao}, \bibinfo{person}{Sarah Wiegreffe}, \bibinfo{person}{Uri Alon}, \bibinfo{person}{Nouha Dziri}, \bibinfo{person}{Shrimai Prabhumoye}, \bibinfo{person}{Yiming Yang}, \bibinfo{person}{Shashank Gupta}, \bibinfo{person}{Bodhisattwa~Prasad Majumder}, \bibinfo{person}{Katherine Hermann}, \bibinfo{person}{Sean Welleck}, \bibinfo{person}{Amir Yazdanbakhsh}, {and} \bibinfo{person}{Peter Clark}.} \bibinfo{year}{2023}\natexlab{}.
\newblock \showarticletitle{Self-{Refine}: Iterative {Refinement} with {Self}-{Feedback}.}. In \bibinfo{booktitle}{\emph{Conference on {Neural} {Information} {Processing} {Systems} ({NeurIPS})}}.
\newblock


\bibitem[Prajwal et~al\mbox{.}(2019)]%
        {prajwal2019towards}
\bibfield{author}{\bibinfo{person}{KR Prajwal}, \bibinfo{person}{CV Jawahar}, {and} \bibinfo{person}{Ponnurangam Kumaraguru}.} \bibinfo{year}{2019}\natexlab{}.
\newblock \showarticletitle{Towards increased accessibility of meme images with the help of rich face emotion captions}. In \bibinfo{booktitle}{\emph{ACM {International} {Conference} on {Multimedia} ({MM})}}. \bibinfo{pages}{202--210}.
\newblock


\bibitem[Pramanick et~al\mbox{.}(2021a)]%
        {pramanick2021detecting}
\bibfield{author}{\bibinfo{person}{Shraman Pramanick}, \bibinfo{person}{Dimitar Dimitrov}, \bibinfo{person}{Rituparna Mukherjee}, \bibinfo{person}{Shivam Sharma}, \bibinfo{person}{Md.~Shad Akhtar}, \bibinfo{person}{Preslav Nakov}, {and} \bibinfo{person}{Tanmoy Chakraborty}.} \bibinfo{year}{2021}\natexlab{a}.
\newblock \showarticletitle{Detecting {Harmful} {Memes} and {Their} {Targets}.}. In \bibinfo{booktitle}{\emph{Annual {Meeting} of the {Association} for {Computational} {Linguistics} ({ACL})}}. \bibinfo{pages}{2783--2796}.
\newblock


\bibitem[Pramanick et~al\mbox{.}(2021b)]%
        {pramanick2021momenta}
\bibfield{author}{\bibinfo{person}{Shraman Pramanick}, \bibinfo{person}{Shivam Sharma}, \bibinfo{person}{Dimitar Dimitrov}, \bibinfo{person}{Md.~Shad Akhtar}, \bibinfo{person}{Preslav Nakov}, {and} \bibinfo{person}{Tanmoy Chakraborty}.} \bibinfo{year}{2021}\natexlab{b}.
\newblock \showarticletitle{MOMENTA: A {Multimodal} {Framework} for {Detecting} {Harmful} {Memes} and {Their} {Targets}.}. In \bibinfo{booktitle}{\emph{Conference on {Empirical} {Methods} in {Natural} {Language} {Processing} ({EMNLP})}}. \bibinfo{pages}{4439--4455}.
\newblock


\bibitem[Radford et~al\mbox{.}(2021)]%
        {radford2021learning}
\bibfield{author}{\bibinfo{person}{Alec Radford}, \bibinfo{person}{Jong~Wook Kim}, \bibinfo{person}{Chris Hallacy}, \bibinfo{person}{Aditya Ramesh}, \bibinfo{person}{Gabriel Goh}, \bibinfo{person}{Sandhini Agarwal}, \bibinfo{person}{Girish Sastry}, \bibinfo{person}{Amanda Askell}, \bibinfo{person}{Pamela Mishkin}, \bibinfo{person}{Jack Clark}, {et~al\mbox{.}}} \bibinfo{year}{2021}\natexlab{}.
\newblock \showarticletitle{Learning transferable visual models from natural language supervision}. In \bibinfo{booktitle}{\emph{International Conference on Machine Learning (ICML)}}. PMLR, \bibinfo{pages}{8748--8763}.
\newblock


\bibitem[Stolfo et~al\mbox{.}(2024)]%
        {stolfo2024confidence}
\bibfield{author}{\bibinfo{person}{Alessandro Stolfo}, \bibinfo{person}{Ben Wu}, \bibinfo{person}{Wes Gurnee}, \bibinfo{person}{Yonatan Belinkov}, \bibinfo{person}{Xingyi Song}, \bibinfo{person}{Mrinmaya Sachan}, {and} \bibinfo{person}{Neel Nanda}.} \bibinfo{year}{2024}\natexlab{}.
\newblock \showarticletitle{Confidence regulation neurons in language models}.
\newblock \bibinfo{journal}{\emph{Conference on {Neural} {Information} {Processing} {Systems} ({NeurIPS})}}  \bibinfo{volume}{37} (\bibinfo{year}{2024}).
\newblock


\bibitem[Sturua et~al\mbox{.}(2024)]%
        {sturua2024jina}
\bibfield{author}{\bibinfo{person}{Saba Sturua}, \bibinfo{person}{Isabelle Mohr}, \bibinfo{person}{Mohammad~Kalim Akram}, \bibinfo{person}{Michael G{\"u}nther}, \bibinfo{person}{Bo Wang}, \bibinfo{person}{Markus Krimmel}, \bibinfo{person}{Feng Wang}, \bibinfo{person}{Georgios Mastrapas}, \bibinfo{person}{Andreas Koukounas}, \bibinfo{person}{Nan Wang}, {et~al\mbox{.}}} \bibinfo{year}{2024}\natexlab{}.
\newblock \showarticletitle{jina-embeddings-v3: Multilingual embeddings with task lora}.
\newblock \bibinfo{journal}{\emph{arXiv preprint arXiv:2409.10173}} (\bibinfo{year}{2024}).
\newblock


\bibitem[Thrun and Pratt(1998)]%
        {thrun1998learning}
\bibfield{author}{\bibinfo{person}{Sebastian Thrun} {and} \bibinfo{person}{Lorien Pratt}.} \bibinfo{year}{1998}\natexlab{}.
\newblock \showarticletitle{Learning to learn: Introduction and overview}.
\newblock In \bibinfo{booktitle}{\emph{Learning to learn}}. \bibinfo{publisher}{Springer}, \bibinfo{pages}{3--17}.
\newblock


\bibitem[Tian et~al\mbox{.}(2024)]%
        {tian2024learning}
\bibfield{author}{\bibinfo{person}{Yuanhe Tian}, \bibinfo{person}{Fei Xia}, {and} \bibinfo{person}{Yan Song}.} \bibinfo{year}{2024}\natexlab{}.
\newblock \showarticletitle{Learning {Multimodal} {Contrast} with {Cross}-modal {Memory} and {Reinforced} {Contrast} {Recognition}}. In \bibinfo{booktitle}{\emph{Findings of the {Association} for {Computational} {Linguistics} {ACL}}}. Association for Computational Linguistics, \bibinfo{pages}{6561--6573}.
\newblock


\bibitem[Tworkowski et~al\mbox{.}(2023)]%
        {tworkowski2023focused}
\bibfield{author}{\bibinfo{person}{Szymon Tworkowski}, \bibinfo{person}{Konrad Staniszewski}, \bibinfo{person}{Miko{\l}aj Pacek}, \bibinfo{person}{Yuhuai Wu}, \bibinfo{person}{Henryk Michalewski}, {and} \bibinfo{person}{Piotr Mi{\l}o{\'s}}.} \bibinfo{year}{2023}\natexlab{}.
\newblock \showarticletitle{Focused transformer: Contrastive training for context scaling}.
\newblock \bibinfo{journal}{\emph{Conference on {Neural} {Information} {Processing} {Systems} ({NeurIPS})}}  \bibinfo{volume}{36} (\bibinfo{year}{2023}), \bibinfo{pages}{42661--42688}.
\newblock


\bibitem[Wang et~al\mbox{.}(2021)]%
        {wang2021tent}
\bibfield{author}{\bibinfo{person}{Dequan Wang}, \bibinfo{person}{Evan Shelhamer}, \bibinfo{person}{Shaoteng Liu}, \bibinfo{person}{Bruno~A. Olshausen}, {and} \bibinfo{person}{Trevor Darrell}.} \bibinfo{year}{2021}\natexlab{}.
\newblock \showarticletitle{Tent: Fully {Test}-{Time} {Adaptation} by {Entropy} {Minimization}.}. In \bibinfo{booktitle}{\emph{International {Conference} on {Learning} {Representations} ({ICLR})}}.
\newblock


\bibitem[Wei et~al\mbox{.}(2022)]%
        {wei2022chainofthought}
\bibfield{author}{\bibinfo{person}{Jason Wei}, \bibinfo{person}{Xuezhi Wang}, \bibinfo{person}{Dale Schuurmans}, \bibinfo{person}{Maarten Bosma}, \bibinfo{person}{Brian Ichter}, \bibinfo{person}{Fei Xia}, \bibinfo{person}{Ed~H. Chi}, \bibinfo{person}{Quoc~V. Le}, {and} \bibinfo{person}{Denny Zhou}.} \bibinfo{year}{2022}\natexlab{}.
\newblock \showarticletitle{Chain-of-{Thought} {Prompting} {Elicits} {Reasoning} in {Large} {Language} {Models}.}. In \bibinfo{booktitle}{\emph{Conference on {Neural} {Information} {Processing} {Systems} ({NeurIPS})}}.
\newblock


\bibitem[Xu et~al\mbox{.}(2025)]%
        {xu2025hyperhateprompt}
\bibfield{author}{\bibinfo{person}{Bo Xu}, \bibinfo{person}{Erchen Yu}, \bibinfo{person}{Jiahui Zhou}, \bibinfo{person}{Hongfei Lin}, {and} \bibinfo{person}{Linlin Zong}.} \bibinfo{year}{2025}\natexlab{}.
\newblock \showarticletitle{HyperHatePrompt: A {Hypergraph}-based {Prompting} {Fusion} {Model} for {Multimodal} {Hate} {Detection}.}. In \bibinfo{booktitle}{\emph{International {Conference} on {Computational} {Linguistics} ({COLING})}}.
\newblock


\bibitem[Xu et~al\mbox{.}(2020)]%
        {xu2020multi}
\bibfield{author}{\bibinfo{person}{Xu Xu}, \bibinfo{person}{Youwei Jia}, \bibinfo{person}{Yan Xu}, \bibinfo{person}{Zhao Xu}, \bibinfo{person}{Songjian Chai}, {and} \bibinfo{person}{Chun~Sing Lai}.} \bibinfo{year}{2020}\natexlab{}.
\newblock \showarticletitle{A multi-agent reinforcement learning-based data-driven method for home energy management}.
\newblock \bibinfo{journal}{\emph{IEEE Transactions on Smart Grid}} \bibinfo{volume}{11}, \bibinfo{number}{4} (\bibinfo{year}{2020}), \bibinfo{pages}{3201--3211}.
\newblock


\bibitem[Yang et~al\mbox{.}(2023)]%
        {yang2023invariant}
\bibfield{author}{\bibinfo{person}{Chuanpeng Yang}, \bibinfo{person}{Fuqing Zhu}, \bibinfo{person}{Jizhong Han}, {and} \bibinfo{person}{Songlin Hu}.} \bibinfo{year}{2023}\natexlab{}.
\newblock \showarticletitle{Invariant {Meets} {Specific}: A {Scalable} {Harmful} {Memes} {Detection} {Framework}.}. In \bibinfo{booktitle}{\emph{ACM {International} {Conference} on {Multimedia} ({MM})}}. \bibinfo{pages}{4788--4797}.
\newblock


\bibitem[Yang et~al\mbox{.}(2022)]%
        {yang2022vision}
\bibfield{author}{\bibinfo{person}{Jinyu Yang}, \bibinfo{person}{Jiali Duan}, \bibinfo{person}{Son Tran}, \bibinfo{person}{Yi Xu}, \bibinfo{person}{Sampath Chanda}, \bibinfo{person}{Liqun Chen}, \bibinfo{person}{Belinda Zeng}, \bibinfo{person}{Trishul Chilimbi}, {and} \bibinfo{person}{Junzhou Huang}.} \bibinfo{year}{2022}\natexlab{}.
\newblock \showarticletitle{Vision-language pre-training with triple contrastive learning}. In \bibinfo{booktitle}{\emph{Proceedings of the IEEE/CVF Conference on Computer Vision and Pattern Recognition (CVPR)}}. \bibinfo{pages}{15671--15680}.
\newblock


\bibitem[Zhang et~al\mbox{.}(2022)]%
        {zhang2022opt}
\bibfield{author}{\bibinfo{person}{Susan Zhang}, \bibinfo{person}{Stephen Roller}, \bibinfo{person}{Naman Goyal}, \bibinfo{person}{Mikel Artetxe}, \bibinfo{person}{Moya Chen}, \bibinfo{person}{Shuohui Chen}, \bibinfo{person}{Christopher Dewan}, \bibinfo{person}{Mona Diab}, \bibinfo{person}{Xian Li}, \bibinfo{person}{Xi~Victoria Lin}, {et~al\mbox{.}}} \bibinfo{year}{2022}\natexlab{}.
\newblock \showarticletitle{Opt: Open pre-trained transformer language models}.
\newblock \bibinfo{journal}{\emph{arXiv preprint arXiv:2205.01068}} (\bibinfo{year}{2022}).
\newblock


\bibitem[Zhao et~al\mbox{.}(2024)]%
        {zhao2024expel}
\bibfield{author}{\bibinfo{person}{Andrew Zhao}, \bibinfo{person}{Daniel Huang}, \bibinfo{person}{Quentin Xu}, \bibinfo{person}{Matthieu Lin}, \bibinfo{person}{Yong-Jin Liu}, {and} \bibinfo{person}{Gao Huang}.} \bibinfo{year}{2024}\natexlab{}.
\newblock \showarticletitle{ExpeL: LLM {Agents} {Are} {Experiential} {Learners}.}. In \bibinfo{booktitle}{\emph{AAAI {Conference} on {Artificial} {Intelligence} ({AAAI})}}. \bibinfo{pages}{19632--19642}.
\newblock


\end{thebibliography}

\newpage

\appendix

\section{Detailed Prompt Designing for \M}
\label{app-prompt}
In this section, we provide the designing details and the content (example) of each prompt in \M.

\subsection{Prompt for Explicit Meme Identification}
To design a prompt for accurately isolating explicit memes from the original dataset, the first is to instruct the LMM to determine whether a given meme contains harmful content. 
Additionally, it is crucial to address overconfident predictions, as this may result in the misidentification of complex memes as explicit, thereby introducing an excessive number of erroneously pseudo-labeled memes.
Based on these considerations, we design the following prompt:

\begin{tcolorbox}[colback=gray!10, colframe=black, width=0.49\textwidth, boxrule=0.1mm, title=\textbf{Prompt for Explicit Meme Identification.}]
    
    \textbf{{Prompt:}} {Given the meme with its image and the textual element embedded in the image, your task is to carefully and critically assess whether this meme is harmful or not, in order to maintain the benignness and integrity of information on the Internet. 
    Please leverage your extensive knowledge to deeply analyze and understand this meme, and give your final judgment.
    
    Please note that while your primary goal is to provide a judgment after thoughtful analysis, it’s important to avoid overgeneralizing or being overly conclusive in cases where ambiguity exists.
    You must only return `1' for harmful, or `0' for benign.
    }

    \textbf{{Image:}} { \{\oc{\textbf{Meme Image}}\}}
    
    \textbf{{Textual Element:}} { \{\oc{\textbf{Meme Textual Element}}\}}

\end{tcolorbox}

\subsection{Prompt for Experience Gathering}
To gather diverse experiences while minimizing the hallucinations, we design a two-step Chain-of-Thought~\cite{wei2022chainofthought} (CoT) based prompt. 
It enhances the reliability of the experience by first guiding the LMM agent to describe the content of the two memes, and analyzing their differences based on the content. 
This prompt is shown below:

\begin{tcolorbox}[colback=gray!10, colframe=black, width=0.49\textwidth, boxrule=0.2mm, title=\textbf{Prompt for Experience Gathering.}]
    
    \textbf{{Prompt:}} {Given two memes that are visually or structurally similar but belong to distinct categories: Meme i, which is harmful, and Meme j, which is benign. Please complete the following two steps:
    
    Step 1: 
   Clearly summarize the content of each meme by carefully analyzing its image and textual element embedded in the image, and considering any implicit or explicit messages it conveys.
   
    Step 2: 
      Based on the content of two memes, contrast the key differences between them to explain why Meme i is classified as harmful content, while Meme j remains benign.}

    \textbf{{Image of Meme i:}} { \{\oc{\textbf{Meme Image i}}\}}
    
    \textbf{{Textual Element of Meme i:}} { \{\oc{\textbf{Meme Textual Element i}}\}}

     \textbf{{Image of Meme j:}} { \{\oc{\textbf{Meme Image j}}\}}
    
     \textbf{{Textual Element of Meme j:}} { \{\oc{\textbf{Meme Textual Element j}}\}}
    
\end{tcolorbox}

\subsection{Prompt for Reference Refinement}
In the reference refinement process, we design a prompt that endows the LMM agent with four atomic operations—ADD, UPVOTE, DOWNVOTE, and EDIT—which the agent can use to iteratively update the reference set based on each new experience.
The content of this prompt is presented below:

\begin{tcolorbox}[colback=gray!10, colframe=black, width=0.485\textwidth, boxrule=0.2mm, title=\textbf{Prompt for Reference Refinement.}]
    
    \textbf{{Prompt:}} {You have a set of experiences for identifying harmful memes, originally created by comparing similar but contradictory categories of two memes.
    Now, a new experience arrives containing the description of one harmful and one similar but benign one, and a summary of the differences between the them.
    Your task is to distill new references from the experience and update the existing references by choosing one operation: : ADD, EDIT, UPVOTE, and DOWNVOTE.

Strict Rules:

1. ADD only if:
   - add new references that are very different from exisitng references and relevatnt for other detection.
   
2. EDIT must:
   - if any existing reference is not general enough or can be enhanced, rewrite and improve it.
   
3. UPVOTE if:
   - if the existing reference is strongly relevant for current reference
   
4. DOWNVOTE if:
   - if one exisiing reference is contradictory or similar/duplicated to other existing reference. 
   
5. Maximum {size} references preserved

6. Output only valid JSON

Context:

Current references Set (importance order):
\{cur\_set\_str\}

New Coming Experience:

\{new\_experience\}

Processing Steps:

1. Ensure the added and edited references are concise, clear while keeping them 2 or 3 sentences.

2. Ensure the references are concise and easy to follow.

3. Actively downvote references that are vague or hard to understand, and maintain the reference set at {size} items.

4. Try to make every reference useful, make more upvotes, and downvotes.

5. Refine references to emphasize distinct signals that uniquely identify specific harmful patterns.

6. Generalize references to extract universal principles that capture common traits of harmful content.

7. Return only the JSON operations with the below format:

[
  {{
    \texttt{"operation": "<ADD|EDIT|UPVOTE|DOWNVOTE>"},
    
    \texttt{"target": "<index/none>"},
    
    \texttt{"reference": "<new/revised text>"}
  }}
]
}

    \textbf{{Reference Set:}} { \{\oc{\textbf{Current Reference Set}}\}}
    
    \textbf{{Experience:}} { \{\oc{\textbf{Current Experience}}\}}

\end{tcolorbox}

\subsection{Prompting LMM Agent for Inference}
During the inference, we design a prompt to guide the LMM agent for classification of a given meme by leveraging the derived references. The content of this prompt is presented below:

\begin{tcolorbox}[colback=gray!10, colframe=black, width=0.485\textwidth, boxrule=0.2mm, title=\textbf{Prompt for Inference.}]
    
    \textbf{{Prompt:}} {
    Given a meme with its image and textual element embedded in the image, your task is to determine whether this meme is harmful or benign by considering its multimodal content.
    Moreover, a set of valuable detection references are also provided to serve as additional knowledge, which will help you in determine the class of the given meme.
    Please leverage your pre-trained knowledge and the references to analyze and understand this meme, and give your final judgment.
    Your output should strictly follow the format: Thought: [Your analysis] Answer: [harmful/benign].}

    \textbf{{Image:}} { \{\oc{\textbf{Meme Image}}\}}
    
    \textbf{{Textual Element:}} { \{\oc{\textbf{Meme Textual Element}}\}}

    \textbf{{Reference Set:}} { \{\oc{\textbf{Final Reference Set}}\}}

\end{tcolorbox}

\section{Detailed Experimental Settings}
\label{app-exp-set}

\subsection{Datasets}
In this study, to comprehensively evaluate the effectiveness of our \M, we select three diverse benchmarks, including two widely adopted English meme datasets FHM~\cite{kiela2020hateful}, MAMI~\cite{fersini2022semeval}, and a Chinese meme dataset ToxiCN~\cite{lu2024towards}.
\begin{itemize} [leftmargin=*, topsep=2pt, partopsep=0pt]
\item \textbf{FHM}: The FHM dataset was constructed and released by Facebook as part of a challenge to crowdsource multimodal hateful meme classification solutions.
It comprises 10,000 memes, constructed by reconstructing in-the-wild memes using licensed stock images from Getty Images to ensure legal compliance. 

\item \textbf{MAMI}: The MAMI dataset is a labeled benchmark collection of memes developed for the task of misogynous meme identification. 
It contains a total of 11,000 memes, curated from popular social media platforms (e.g., Twitter and Reddit) and meme-specific websites (e.g., 9GaG, Knowyourmeme, and
Imgur) using methods such as site scraping, manual downloads, and keyword/hashtag-based searches. 

\item \textbf{ToxiCN}: ToxiCN is the first comprehensive dataset focused on harmful memes in Chinese, comprising 12,000 samples sourced from various platforms such as Weibo and Baidu Tieba, covering multiple types of harmful content.

\end{itemize}

\subsection{Baselines}
To comprehensively evaluate the effectiveness of \M, we compare it with \NumBaseline baselines, which can be divided into two groups:
(1) \textit{Label-Driven Methods} which require annotated training data for optimizing the models, including MOMENTA~\cite{pramanick2021momenta}, PromptHate~\cite{cao2022prompting}, MR.HARM~\cite{lin2023beneath}, Pro-Cap~\cite{cao2023procap}, ISM~\cite{yang2023invariant}, ExplainHM~\cite{lin2024towards}, and HHPrompt~\cite{xu2025hyperhateprompt}.
(2) \textit{Few-Shot Learning Methods} which leverage only a few labeled samples to enhance the detection capability, including OPT-30B~\cite{zhang2022opt}, OpenFlamingo-9B~\cite{alayrac2022flamingo}, Qwen2.5-VL-72B~\cite{bai2025qwen2}, GPT-4o~\cite{openai2024gpt}, Mod-HATE~\cite{cao2024modularized}, and LoReHM~\cite{huang2024towards}.

\begin{table}[t]
\caption{Prompt for the few-shot LMM baselines.}
\vspace{-3mm}
\label{tab:prompt-few-shot-baseline}
\centering
\begin{tabular}{p{0.95\linewidth}}
\Xhline{1pt}
\textbf{Prompt}: 
Given a meme with its image and textual element embedded in the image, your task is to determine whether this meme is harmful or benign by considering its multimodal content.
I will provide you with a list of meme examples along with their detection results. Each meme will include its caption and the text elements within it.
Please leverage your pre-trained knowledge and the examples to analyze and understand this meme, and give your final judgment. Your output should strictly follow the format:

Thought: [Your analysis]
Answer: [harmful/benign]
\\
\textbf{Image}: \{Meme Image\}\\
\textbf{Textual Element}: \{Meme Textual Element\}\\
\textbf{Examples}: Meme i, \{Meme Caption\}, \{Meme Textual Element\}, \{Prediction Result\}...\\
\Xhline{1pt}
\end{tabular}
\end{table}

\subsection{Implementation Details}
We present the implementations details of this work, including the LMM backbone, data preprocessing, hyper-parameter, baseline implementation, and running environment.
\begin{itemize} [leftmargin=*]
\item \textbf{LMM Backbone:} In this work, we mainly adopt Qwen2.5-VL-72B model --- the latest and powerful open-source model in the Qwen vision-language series --- as the LMM backbone.
Moreover, as \M has strong scalability and can also applied to any close-source LMMs, we also select GPT-4o (gpt-4o-2024-11-20) as another backbone.
To stabilize inference, the temperature of LMMs is set at 0 by default.
For Qwen2.5-VL-72B, we deploy it locally, while for GPT-4o, we access it via the official API service.
Notably, all the open-source LMMs used in this work are instruction-tuned, indicating that the models have been aligned to follow human instructions.

\item \textbf{Data Preprocessing:} 
The meme image and text for each dataset are provided by their original papers.
For the FHM dataset, we also employ the preprocessed text from the prior work~\cite{cao2022prompting}.
During the multimodal retrieval, we leverage the text and vision encoders from the pre-trained CLIP model. 
Specifically, we adopt the \texttt{jina-clip-v2}, a pre-trained variant of CLIP that integrates Jina-XLM-RoBERTa~\cite{sturua2024jina} for text encoding and EVA02~\cite{fang2024eva} family of ViT models for vision encoding.
The meme image and textual element are first input to the vision and text encoders, respectively. And the [CLS] tokens from both encoders are then leveraged to calculate the cosine similarity for retrieval.

\item \textbf{Hyper-Parameter:} There are two key hyper-parameters in \M: the confidence selection ratio $\tau$ and the reference set capacity $N$. 
The value of ratio $\tau$ is selected from the [0-1] and the optimal performances are reached when $\tau = 0.2$ across three datasets. 
The value of capacity $N$ is chosen from the [1-25] and the best results are obtained when $L = 15$ on three datasets.

\item \textbf{Baseline Implementation:} For baseline models, we primarily report the performance results directly from their original papers. 
In cases where these baselines were not previously evaluated on our selected benchmarks or adopted different metrics, we strictly follow their original experimental settings to faithfully reproduce their reported results. 
Additionally, for in-context few-shot learning baseline models, we follow the prior work~\cite{huang2024towards} to provide them with 50 textual in-context demonstrations.
And we provide the prompt for the few-shot LMM baselines in \Cref{tab:prompt-few-shot-baseline}.

\item \textbf{Running Environment:}  All experiments are conducted on a system equipped with an AMD EPYC 7K62 CPU, an NVIDIA L40s GPU, and 128 GB of system RAM.

\end{itemize}

\end{document}